\documentclass[conference]{IEEEtran}
\usepackage{amssymb}
\usepackage{makecell}
\usepackage[table]{xcolor}
\usepackage{tabularx}
\usepackage[numbers]{natbib}
\usepackage{times}
\usepackage{amsmath}
\usepackage{amssymb}
\usepackage{siunitx}
\usepackage{makecell}
\usepackage{array}
\usepackage[numbers]{natbib}
\usepackage{multicol}
\usepackage{graphicx}
\definecolor{linkblue}{HTML}{1F4E79}
\definecolor{citegreen}{HTML}{17643B}
\definecolor{urlbrown}{HTML}{924500}

\usepackage[bookmarks=true]{hyperref}

\hypersetup{
    colorlinks=true,
    linkcolor=linkblue,
    citecolor=citegreen,
    urlcolor=urlbrown,
    pdfborder={0 0 0},
    pdftitle={MIKASA-Robo-MMM: Evaluating Memory in Mobile Manipulation}
}

\makeatletter
\let\mmm@maketitle\@maketitle
\renewcommand{\@maketitle}{%
  \mmm@maketitle
  \begin{center}
    \includegraphics[width=\textwidth]{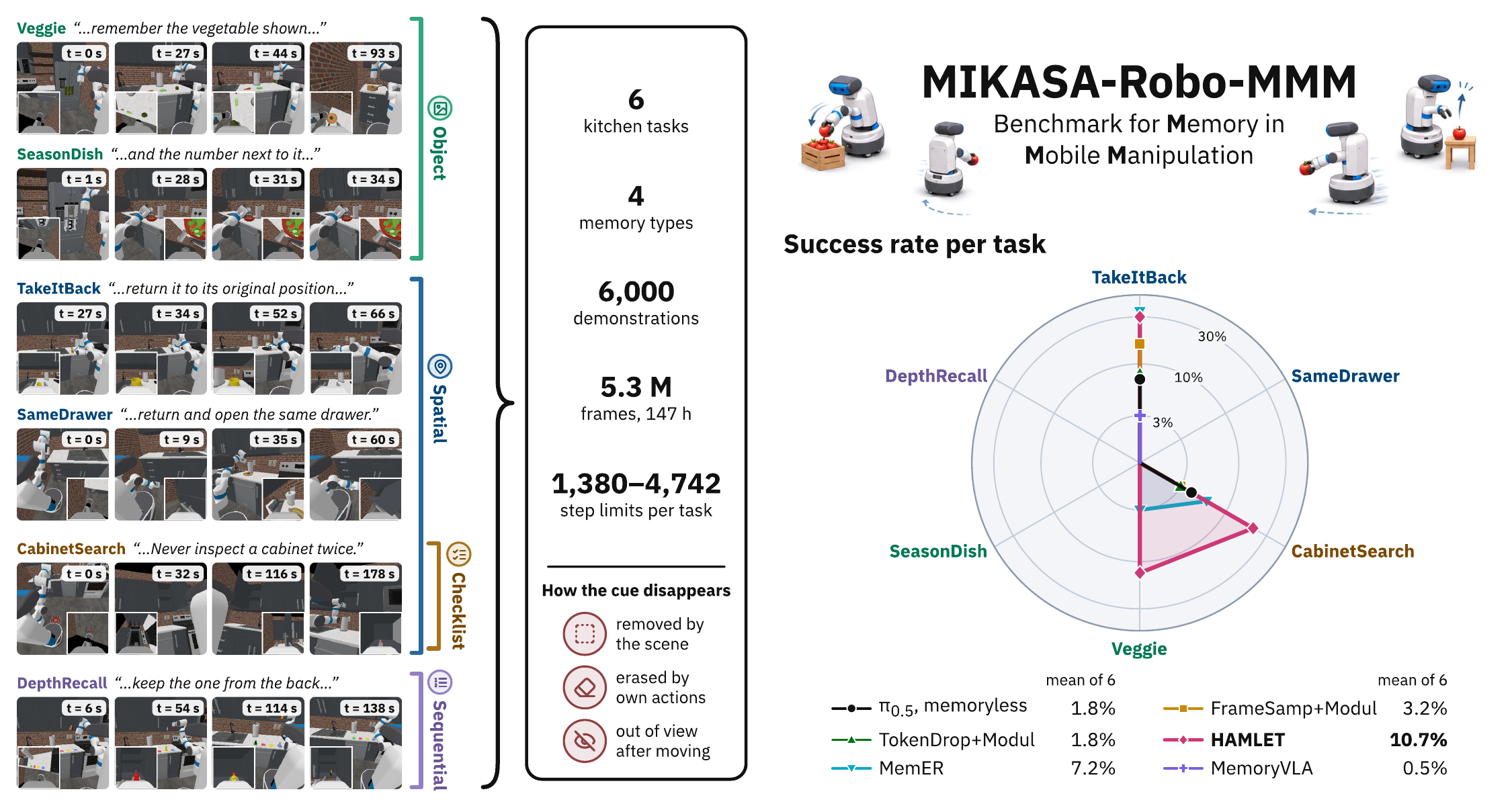}
  \end{center}
  {\def\@captype{figure}%
    \vspace{-20pt}%
    \caption{\textbf{MIKASA-Robo-MMM, a benchmark for evaluating memory in mobile manipulation.}
  The six simulated kitchen tasks shown at left require the robot to act on
  a visual cue after it has disappeared, for example when the robot turns
  away from an object or closes a drawer. Policies must make these decisions
  using head and wrist camera views and proprioception, without access to
  the global pose of the mobile base. The colored brackets mark object,
  spatial, checklist, and sequential memory requirements, with
  \textsc{CabinetSearch} spanning both spatial and checklist memory.
  We fine-tuned five memory-augmented vision-language-action policies and a
  memoryless $\pi_{0.5}$ control on 1{,}000 planner-generated demonstrations
  per task, then evaluated all six on the same 100 held-out seeds for each task.
  In the comparison at right, HAMLET and MemER achieved the highest mean
  success rates (10.7\% and 7.2\%, respectively, versus 1.8\% for the control).
  Three of the six tasks remain unsolved by all evaluated policies, highlighting the need for more effective memory mechanisms in long-horizon mobile manipulation.}

    \label{fig:teaser}%
  }
    \vspace{-10pt}%
}
\makeatother

\newcommand{\taskfield}[1]{\par\smallskip\noindent\textbf{#1.}~}

\begin{document}
\title{MIKASA-Robo-MMM: Evaluating Memory in Mobile Manipulation}
\author{Author Names Omitted for Anonymous Review. Paper-ID [add your ID here]}
%

\author{
    \authorblockN{
        Georgii Volovikov\textsuperscript{1,2*},
        Dair Nurtayev\textsuperscript{2*},
        Egor Cherepanov\textsuperscript{2,3},
        Aleksandr I. Panov\textsuperscript{2,3},
        Alexey K. Kovalev\textsuperscript{2,3}
    }

    \authorblockA{
        \textsuperscript{1}Irkutsk State University,
        \textsuperscript{2}MIRIAI,
        \textsuperscript{3}Cognitive AI Systems Lab
    }

    \authorblockA{
        \textsuperscript{*}Equal contribution
        \\[3pt]
        \href{https://mikasa-robo-mmm.github.io}{
            \textbf{mikasa-robo-mmm.github.io}
        }
    }
}
\maketitle
\begin{abstract}
A household robot moving between work areas often needs to remember what it
saw in one place to decide what to do in another. Many vision-language-action
(VLA) policies use only the current observation, and evaluations of
memory-augmented VLAs have focused largely on tabletop manipulation.
We introduce MIKASA-Robo-MMM (Memory in Mobile Manipulation), a benchmark
that tests whether a robot can use an earlier observation to guide a later
action as it moves through a simulated kitchen. In one task, for example,
the robot must remember which of four drawers it closed so it can reopen
the same one after putting an apple on a plate at the far end of the counter.
We generated 6{,}000 demonstrations with task-specific motion planners and
used them to fine-tune five VLAs with different memory mechanisms and a
memoryless VLA that served as the control.
HAMLET, the strongest memory-augmented policy, achieved a mean success rate
of 10.7\%, compared with 1.8\% for the control. HAMLET outperformed the
control on three tasks, while every policy failed all 100 trials on each
of the other three.
These tasks motivate new memory mechanisms for mobile manipulation,
particularly for remembering the order of objects or how many times to
repeat an action after a visual cue disappears.
Code, data, and videos are available at
\href{https://mikasa-robo-mmm.github.io}{mikasa-robo-mmm.github.io}.
\end{abstract}
\IEEEpeerreviewmaketitle

\section{Introduction}
A kitchen robot is asked to move a cup from the counter onto a tray and
then put it back where it was (Fig.~\ref{fig:teaser}). Once the cup is on
the tray, nothing marks its original spot on the counter, so a
vision-language-action (VLA) policy~\citep{rt2,openvla,pi05} must remember
where the cup stood to complete the task.

Benchmarks for memory-intensive tabletop manipulation test whether a robot
can recall an object's identity, position, or place in a sequence after
the visual cue disappears~\citep{mikasa,mikasavla,robomme,robodojo,liberomem}.
On a tabletop, the arm can return to a remembered location from a fixed
base. In navigation, a robot with neither a map nor a known pose must
relate its current view to landmarks it remembers to track its position
and find its way back. In GOAT-Bench~\citep{goatbench}, for example,
remembering where the robot saw an object on the way to one goal can
shorten its route when that object becomes the next goal.

Mobile manipulation combines these demands because the robot must remember
both how to return to a work area and what it did there. In household
benchmarks such as RoboCasa~\citep{robocasa}, RoboCasa365~\citep{robocasa365},
ManiSkill-HAB~\citep{mshab}, and BEHAVIOR-1K~\citep{behavior1k}, the need for
memory can arise implicitly when the robot turns away from an object it
must handle later.

We introduce \textbf{MIKASA-Robo-MMM} (\textbf{M}emory in \textbf{M}obile
\textbf{M}anipulation) to compare how VLA policies use earlier observations
across six memory-intensive tasks, each designed around a later decision
that depends on information the robot saw earlier. Each task requires the robot to move
its base, taking objects out of view as it turns or drives to another
part of the kitchen. The policy
receives a language instruction, images from head and wrist cameras, and
proprioceptive measurements. Without a kitchen map or the global pose of
the base, it must use its own observations to keep track of the locations
of relevant objects and work areas.

We implement the tasks in ManiSkill3~\citep{maniskill3} using DS-Fetch,
a variant of the Fetch mobile manipulator~\citep{fetch,robobenchmart}, in a simulated RoboCasa
kitchen~\citep{robocasa}. The robot handles household objects such as cups,
vegetables, and condiment bottles and interacts with articulated fixtures,
including hinged cabinet doors and sliding drawers. Depending on the task,
the evaluation horizon ranges from 1{,}380 to 4{,}742 environment steps
at the \qty{10}{\hertz} evaluation interface (approximately 138--474 seconds).
For training, we provide 1{,}000 successful planner-generated demonstrations
per task, averaging 88.5 seconds per demonstration.

The tasks draw on four types of memory. Spatial and object memory allow
the robot to recall where an object stood and which object a cue identified,
respectively. Sequential memory is needed to recover an ordering, such as
the original arrangement of condiments on a shelf, whereas checklist memory
lets the robot keep track of the cabinets it has already inspected during
a search.

\textbf{Our main contributions are as follows.}
\begin{enumerate}
  \item We introduce \textbf{MIKASA-Robo-MMM}, a benchmark for memory in
    mobile manipulation that specifies, for each task, what the robot must
    remember, when the cue disappears, and which later decision depends on it.
  \item We release \textbf{6{,}000 planner-generated demonstrations}
    in LeRobotDataset v3.0 format, with 1{,}000 demonstrations and 100
    separate evaluation seeds per task (Section~\ref{sec:dataset}).
  \item We evaluate \textbf{five memory-augmented VLA policies} alongside a
    memoryless control (Section~\ref{sec:baselines}). HAMLET achieves the
    highest mean success rate (\textbf{10.7\%}, compared with 1.8\% for the
    control), although none of the evaluated policies completes
    \textsc{SameDrawer}, \textsc{SeasonDish}, or \textsc{DepthRecall} in
    100 trials per policy on each task.
\end{enumerate}
\begin{figure*}[t]
  \centering
    \includegraphics[width=\textwidth]{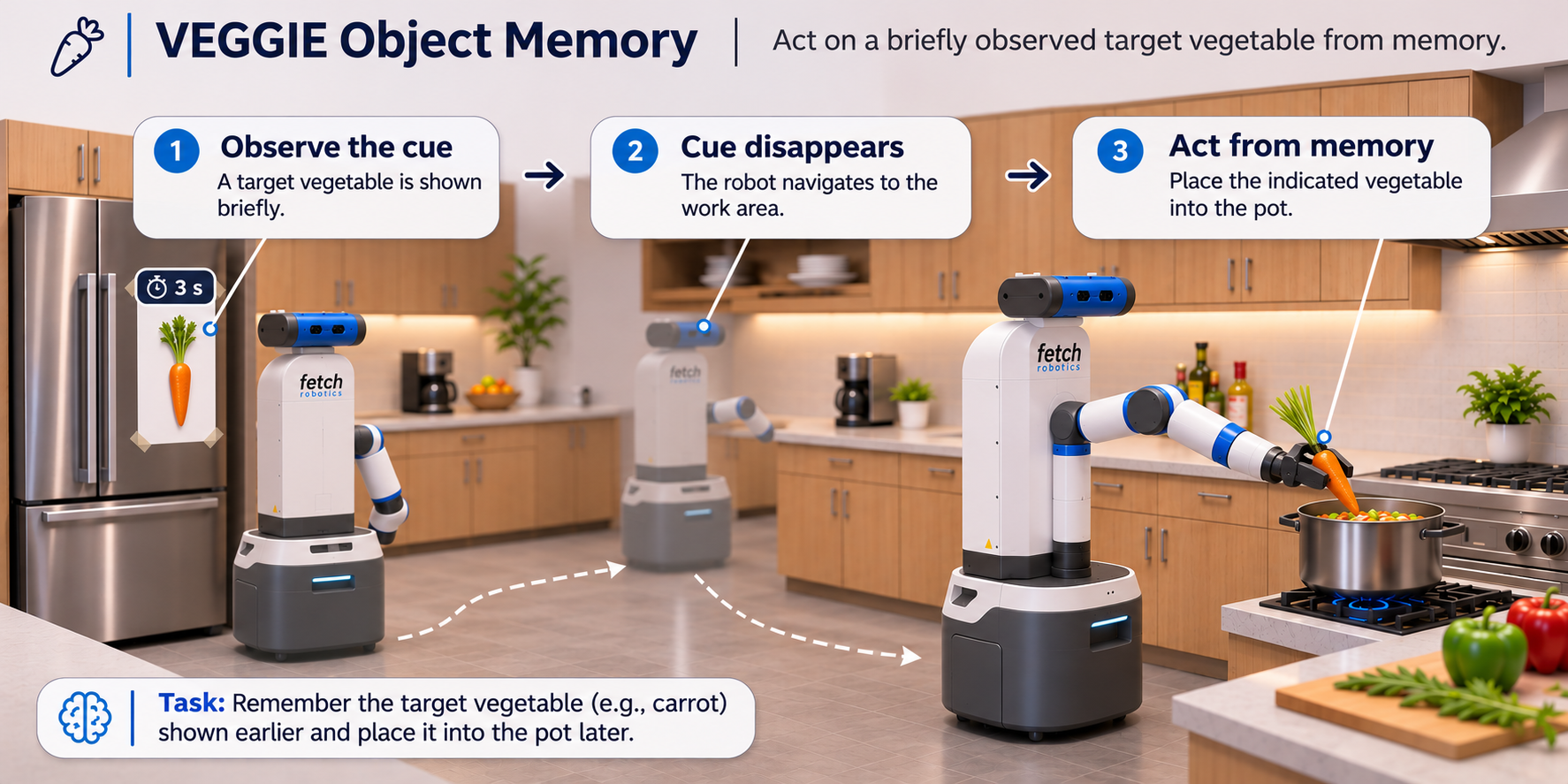}
  \caption{\textbf{Visual abstract.} A mobile manipulator has to act on what it saw
  earlier. In \textsc{Veggie} the target vegetable is shown as a picture taped
  to the fridge door for a few seconds. Once the picture is removed, nothing
  marks which vegetable is meant, and the robot has to carry that cue through
  navigation and manipulation into the pot.}
  \label{fig:visual-abstract}
\end{figure*}

\section{Related Work}
\subsection{\textbf{Memory-Augmented Vision-Language-Action Models}}

A VLA can condition on recent frames, but retaining a longer history
requires deciding what to keep. RoboMME~\citep{robomme} samples frames
across the episode or selects tokens based on visual change, while
HAMLET~\citep{hamlet} compresses observations into learned tokens.
Frame sampling retains the context of an entire image at selected
times, whereas token selection and compression preserve less of each
image in exchange for a more compact representation.
ReMem-VLA~\citep{rememvla} carries recurrent state across frames and
action chunks. When a decision calls for a specific past observation,
MemER~\citep{memer} retrieves keyframes for subgoal prediction, while
MemoryVLA~\citep{memoryvla} retrieves features from a bank and consolidates
similar entries to keep the bank bounded. SOMA~\citep{pmlr-v306-li26ay} aggregates observations from a movable head
camera into a spatial memory that the policy can query when the target
is outside its current field of view. Sampling can miss a cue visible
for only a few frames, while a recurrent state must preserve the cue
through updates triggered by unrelated observations. MIKASA-Robo-MMM
compares five memory-augmented VLAs and a memoryless control on the
same mobile manipulation episodes, where relevant evidence can include
both a brief visual cue and the results of repeated interactions.

\subsection{\textbf{Memory-Intensive Manipulation Benchmarks}}

MIKASA-Robo~\citep{mikasa} uses tabletop tasks to study memory in
reinforcement learning, and MIKASA-Robo-VLA~\citep{mikasavla} adapts
that design to pretrained language-conditioned policies, including
controls with visible cues. LIBERO-Mem~\citep{liberomem} makes object
selection depend on earlier observations or interactions, while
MemoryBench~\citep{sam2act} tests spatial and action recall. In both
cases, the benchmark specifies which piece of earlier information is
relevant to the task, such as an object's identity or location.
RoboMME~\citep{robomme} covers counting and procedural memory, and
RoboDojo~\citep{robodojo} separates memory from long-horizon execution
in its evaluation. A long sequence can remain reactive if each
intermediate result makes the next action evident. Conversely, even a
short task requires memory when identical observations call for
different actions depending on the history.
MIKASA-Robo-MMM extends these tests to a moving base, so the robot may
have to use a remembered cue after leaving a work area or returning
from a different pose, when a location seen earlier has to be related
to the robot's current workspace.

\subsection{\textbf{Memory in Robot Navigation}}

Without a prior map or known global pose, history helps a robot recognize
places, estimate its position, and plan routes. Goal-Oriented Semantic
Exploration~\citep{semanticexploration} maps observed objects and free
space so that semantic context can guide exploration. Semi-Parametric
Topological Memory~\citep{sptm} instead represents previously seen
places and how to move between them. GOAT~\citep{goat}
retains object instances for later goals, and Scene Memory
Transformer~\citep{scenememorytransformer} learns to attend to earlier
observations. These representations support different decisions, from
choosing an unexplored region to returning to a specific object seen
earlier. GOAT-Bench~\citep{goatbench} tests reuse of history across
successive goals, where remembering what the robot saw on one leg can
help it find a later target. MessyMem~\citep{banwasi2026messymem} carries the outcomes of manipulation
across tasks, for example by recording that a cabinet is locked. In MIKASA-Robo-MMM, a successful return can
still leave a decision that depends on history, such as which drawer
to open or where to place a cup.

\subsection{\textbf{Mobile Manipulation Benchmarks}}

Composite tasks in RoboCasa~\citep{robocasa}, RoboCasa365~\citep{robocasa365},
and BEHAVIOR-1K~\citep{behavior1k} can create incidental memory demands
when objects leave view during travel or intermediate actions. The
source of partial observability can be the robot's own movement or a
change caused by a subtask, such as putting an object inside a cabinet.
ManiSkill-HAB~\citep{mshab} teleports the robot between subtasks to focus
evaluation on manipulation, removing the intervening sequence of
viewpoints from its long-horizon protocol.

RoboCasa-Mem~\citep{robocasamem} ties later success conditions to earlier
interactions, for example by asking the robot to report how many drawers
it inspected. The delay before recall can vary within a single episode,
and identical delays can test different demands, such as retaining a
single visual cue or updating a record of successive interactions.
RoboCasa-Long from MOST~\citep{most} tests recall of object-state changes
and task progress across extended routines. Table~\ref{tab:related} summarizes
these differences in task design. MIKASA-Robo-MMM specifies the cue
and dependent decision in each task to test how VLAs use history during
continuous mobile manipulation, where success also requires navigation
and motor execution.

\begin{table*}[t]
    \centering
    \caption{\textbf{Comparison of simulated manipulation benchmarks.}
    MIKASA-Robo-MMM combines explicit memory requirements,
    continuous mobile manipulation, and egocentric observations
    without global base localization.}
    \label{tab:related}

    \footnotesize
    \newcommand{\benchyes}{
        \textcolor[HTML]{17643B}{\ensuremath{\checkmark}}
    }
    \newcommand{\benchno}{
        \textcolor[HTML]{9A3838}{\ensuremath{\times}}
    }
    \newcommand{\benchna}{
        \textcolor[HTML]{888888}{--}
    }
    \newcommand{\benchnr}{
        \textcolor[HTML]{888888}{NR}
    }

    \setlength{\tabcolsep}{2.0pt}
    \renewcommand{\arraystretch}{1.35}

    \resizebox{\textwidth}{!}{%
    \begin{tabular}{@{}lcccccccclrrr@{}}

        \hline

        & \multicolumn{3}{c}{\textbf{Robot setting}}
        & \multicolumn{4}{c}{\textbf{Memory requirements}}
        &
        & \multicolumn{3}{c}{\textbf{Training demonstrations}}
        & \textbf{Evaluation}
        \\

        \cline{2-4}
        \cline{5-8}
        \cline{10-12}

        \textbf{Benchmark}
        & \makecell{Mobile\\base}
        & \makecell{Continuous\\navigation}
        & \makecell{Global pose\\withheld}
        & \makecell{Explicit\\memory tasks}
        & \makecell{Exogenous\\cue removal}
        & \makecell{Interaction-induced\\information loss}
        & \makecell{Base-motion-induced\\information loss}
        & \makecell{Control\\interface}
        & Source
        & \makecell{Demos\\per task}
        & \makecell{Mean\\duration (s)}
        & \makecell{Episodes\\per task}
        \\

        \hline

        MIKASA-Robo~\citep{mikasa}
        & \benchno
        & \benchno
        & \benchna
        & \benchyes
        & \benchyes
        & \benchyes
        & \benchno
        & Joint delta
        & Oracle PPO
        & 1{,}000
        & 4.6
        & 100
        \\

        MIKASA-Robo-VLA~\citep{mikasavla}
        & \benchno
        & \benchno
        & \benchna
        & \benchyes
        & \benchyes
        & \benchyes
        & \benchno
        & EE delta
        & MP + PPO
        & 250
        & 28.0
        & 50
        \\

        LIBERO-Mem~\citep{liberomem}
        & \benchno
        & \benchno
        & \benchna
        & \benchyes
        & \benchno
        & \benchyes
        & \benchno
        & EE delta
        & Human
        & 82--100
        & 16.7
        & 20
        \\

        MemoryBench~\citep{sam2act}
        & \benchno
        & \benchno
        & \benchna
        & \benchyes
        & \benchno
        & \benchyes
        & \benchno
        & \makecell{EE keyframes\\+ MP}
        & Scripted
        & 100
        & $\approx$15.5
        & 25
        \\

        RoboMME~\citep{robomme}
        & \benchno
        & \benchno
        & \benchna
        & \benchyes
        & \benchyes
        & \benchyes
        & \benchno
        & Joint / EE
        & MP
        & 100
        & 48.1
        & 50
        \\

        RoboDojo$^{a}$~\citep{robodojo}
        & \benchno
        & \benchno
        & \benchna
        & \benchyes
        & \benchyes
        & \benchyes
        & \benchno
        & Joint / EE
        & \makecell{MP +\\Human}
        & 100
        & 21.9
        & 50
        \\

        \hline

        RoboCasa~\citep{robocasa}
        & \benchyes
        & \benchyes
        & \benchna
        & \benchno
        & \benchna
        & \benchna
        & \benchna
        & \makecell{EE (OSC)\\+ base vel.}
        & \makecell{Human +\\MimicGen}
        & 50
        & 13.6
        & \benchna
        \\

        RoboCasa365~\citep{robocasa365}
        & \benchyes
        & \benchyes
        & \benchna
        & \benchno
        & \benchna
        & \benchna
        & \benchna
        & \makecell{EE (OSC)\\+ base vel.}
        & \makecell{Human +\\MimicGen}
        & 100 / 500
        & 48.5 / 30.0
        & \benchna
        \\

        ManiSkill-HAB$^{b}$~\citep{mshab}
        & \benchyes
        & \benchno
        & \benchna
        & \benchno
        & \benchna
        & \benchna
        & \benchna
        & \makecell{Joint delta\\+ base vel.}
        & RL
        & \benchna
        & \benchna
        & \benchna
        \\

        BEHAVIOR-1K$^{c}$~\citep{behavior1k}
        & \benchyes
        & \benchyes
        & \benchna
        & \benchno
        & \benchna
        & \benchna
        & \benchna
        & Configurable
        & Human
        & 200
        & 351.5
        & \benchna
        \\

        \hline

        RoboCasa-Mem~\citep{robocasamem}
        & \benchyes
        & \benchyes
        & \benchno
        & \benchyes
        & \benchno
        & \benchyes
        & \benchyes
        & \makecell{Joint pos.\\+ base cmd.}
        & MP
        & 200
        & 94.1
        & 50
        \\

        RoboCasa-Long~\citep{most}
        & \benchyes
        & \benchyes
        & \benchno
        & \benchyes
        & \benchno
        & \benchyes
        & \benchyes
        & \makecell{EE\\+ base cmd.}
        & \benchnr
        & 200
        & $\approx$228
        & 30
        \\

        \hline

        \rowcolor[HTML]{EAF3ED}

        \textbf{MIKASA-Robo-MMM}$^{d}$
        & \benchyes
        & \benchyes
        & \benchyes
        & \benchyes
        & \benchyes
        & \benchyes
        & \benchyes
        & \makecell{Joint pos.\\+ base vel.}
        & \textbf{MP}
        & \textbf{1{,}000}
        & 88.5
        & \textbf{100}
        \\

        \hline

    \end{tabular}%
    }

    \par\smallskip

    \begin{minipage}{\textwidth}
        \footnotesize
        \raggedright

        \benchyes/\benchno: presence/absence of a feature,
        \benchna: not applicable, NR: not reported or verified.
        Memory-related criteria refer to explicitly designed
        memory tasks. Exogenous cue removal is independent
        of robot actions. Global pose refers to the mobile-base
        pose in the policy observation. Evaluation counts are
        reported per task and evaluation run.
        LIBERO-Mem duration uses the 961 demonstrations in the released
        training annotations. MemoryBench duration is estimated from
        recorded frame counts and the RLBench time step. RoboCasa duration
        covers the Human-50 atomic-task subset.
        EE: end-effector control,
        OSC: operational space control,
        MP: motion planning,
        RL: reinforcement learning.
        $^{a}$Memory subset.
        $^{b}$Skill-level demonstrations only.
        $^{c}$Challenge dataset.
        $^{d}$Assisted grasping.

    \end{minipage}
\end{table*}

\section{\textbf{Preliminaries}}
\label{sec:background}

\subsection{\textbf{Partially Observable Decision Making}}

We formulate robotic manipulation as a finite-horizon partially
observable Markov decision process (POMDP)~\citep{kaelbling1998},
with state space $\mathcal{S}$, action space $\mathcal{A}$,
observation space $\Omega$, and horizon $T$.
Each episode starts from an initial state $s_0 \sim p_0$.
At each environment step $t=0,\ldots,T-1$, the agent receives
an observation $o_t \sim \mathcal{O}(\cdot \mid s_t)$
and executes an action $a_t$, inducing a state transition
$s_{t+1} \sim \mathcal{T}(\cdot \mid s_t,a_t)$.
While reactive policies $\pi_{\mathrm{reactive}}(a_t \mid o_t)$
condition on the current observation, history-conditioned policies
$\pi_{\mathrm{history}}(a_t \mid h_t)$ additionally exploit the
interaction history $h_t=(o_0,a_0,\ldots,a_{t-1},o_t)$.
Both policy classes may condition on the current timestep.

\subsection{\textbf{Memory-Intensive Tasks}}

Partial observability can give rise to perceptual
aliasing~\citep{whitehead1991,chrisman1992}, whereby distinct
environment states that require different decisions cannot
be distinguished from the current observation.
Memory enables policies to resolve such ambiguity by conditioning
their decisions on previous interactions.
Given the task objective
$J(\pi)=\mathbb{E}_{\tau\sim\pi}[S(\tau)]$,
where $S(\tau)\in\{0,1\}$ denotes episode success,
we characterize memory-intensive tasks by
\[
    J^\star_{\mathrm{history}}
    > J^\star_{\mathrm{reactive}},
    \qquad
    J^\star_c=\sup_{\pi\in\Pi_c}J(\pi),
\]
where $\Pi_c$ denotes the corresponding policy class.
Solving such tasks requires retaining and utilizing
task-relevant historical information, which may include object
identities, spatial locations, event sequences, or the outcomes
of previous interactions.

\subsection{\textbf{Memory-Augmented Vision-Language-Action Policies}}

A vision-language-action (VLA) policy receives a multimodal
observation $o_t=(I_t,q_t,\ell)$, comprising camera views
$I_t=(I_t^1,\ldots,I_t^K)$, proprioceptive measurements $q_t$,
and a language instruction $\ell$.
A memory-augmented VLA generates an action chunk
\[
    A_t=(a_t,\ldots,a_{t+H-1})
    \sim \pi_\theta(\cdot\mid o_t,m_t),
\]
where $H$ is the prediction horizon and $m_t$ represents
historical information maintained across policy queries.
The memory is updated through an architecture-specific
mechanism that incorporates newly acquired observations.
The policy executes the first $K_{\mathrm{exec}}\leq H$
actions before the next query.
By conditioning action generation on both current observations
and stored historical information, memory-augmented VLAs
can address tasks that require information beyond the
current observation.

\subsection{\textbf{Partial Observability in Mobile Manipulation}}

Mobile manipulation introduces additional sources of partial
observability through base motion, changing viewpoints, and
interactions with the environment.
In our setting, the robot observes the scene through
egocentric cameras and proprioceptive measurements, without
access to a global map, global base pose, or privileged
object states.
Task-relevant information acquired in one work area may
therefore remain necessary for subsequent decisions after
the robot has moved to another area, despite changes in its
reference frame and observable workspace.
MIKASA-Robo-MMM exploits these properties to construct
memory-intensive tasks in which successful manipulation
depends on information acquired before navigation or
intermediate interactions.

\section{MIKASA-Robo-MMM}

MIKASA-Robo-MMM is a benchmark of language-conditioned,
memory-intensive mobile manipulation tasks designed to evaluate
the ability of robotic policies to utilize historical information
during interaction with household environments.
The benchmark extends the MIKASA-Robo family~\citep{mikasa,mikasavla}
to a mobile embodiment, introducing additional memory requirements
arising from navigation, changing viewpoints, and interactions
with objects across spatially distributed work areas.
All tasks share a common robot embodiment, observation and action
spaces, and evaluation interface, while varying the information
that must be retained and the decisions that depend on it.

\subsection{\textbf{Benchmark Design Principles}}
\label{sec:design-choices}

The benchmark is designed around the following principles.

\noindent\textbf{Explicit memory requirements.}
Each task introduces a dependency between information acquired
during earlier interactions and subsequent decisions.
Depending on the task, relevant information is provided by
a temporary visual cue, encoded in the initial configuration
of the environment, or accumulated through interaction with
the scene.
The task objectives and success criteria are constructed
to make this information relevant beyond the observation
in which it was originally available.

\noindent\textbf{Multiple sources of partial observability.}
Task-relevant information can become unavailable when the environment
removes a visual cue, when the robot moves an object or closes a drawer,
or when navigation changes the robot's viewpoint.
These mechanisms can coexist within a single episode, requiring
the policy to retain relevant information across both navigation
and manipulation.

\noindent\textbf{Randomized task instances.}
Task-specific randomization introduces variation in object
identities, initial configurations, spatial arrangements,
visual cues, and robot poses across episodes.
In particular, randomizing the information that determines
subsequent decisions encourages policies to condition their
behavior on the observed task instance rather than memorize
a fixed solution.

\noindent\textbf{Household-oriented interactions.}
The tasks incorporate everyday household activities involving
object transport, retrieval, placement, and interaction with
articulated furniture.
Memory requirements are embedded within these activities,
connecting the retention of historical information with
practical manipulation objectives.

\noindent\textbf{Reduced grasping complexity.}
Following the use of assisted grasping in household simulation
benchmarks~\citep{behavior1k}, we employ an assisted grasp
mechanism that maintains the attachment of a grasped object
to the gripper until release.
This design choice reduces the complexity of grasp maintenance
and places greater emphasis on navigation, object selection,
and memory-dependent manipulation decisions.

\subsection{\textbf{Simulation Environment and Robot Embodiment}}

The benchmark is implemented in ManiSkill3~\citep{maniskill3}
using a RoboCasa kitchen environment~\citep{robocasa}.
We employ DS-Fetch, a variant of the Fetch mobile
manipulator~\citep{fetch} introduced in
RoboBenchMart~\citep{robobenchmart}.
The robot combines a mobile base, a seven-degree-of-freedom
manipulator, a parallel-jaw gripper, a vertically actuated
torso, and a pan-tilt head.

This embodiment enables the robot to interact with objects
distributed across the kitchen and to manipulate articulated
fixtures, including cabinets and drawers.
Independent actuation of the base, torso, and head introduces
viewpoint variability and requires the coordination of
navigation, perception, and manipulation within a shared
control interface.

\subsection{\textbf{Observations and Actions}}
\label{sec:obs-actions}

\noindent\textbf{Observation space.}
At each environment step, the policy receives three egocentric
RGB images, proprioceptive measurements, and a language
instruction specifying the task objective.
Two head-mounted cameras provide $256 \times 256$ RGB images
from complementary viewpoints, while a wrist-mounted camera
provides a $128 \times 128$ RGB image of the local manipulation
workspace.

The 12-dimensional proprioceptive vector comprises the
positions of the seven arm joints, two gripper finger joints,
two head joints, and the torso joint.
The policy operates without access to a global map, global
base pose, or privileged object states.
The global base pose is recorded separately for diagnostic
purposes and is excluded from the policy observation.

\noindent\textbf{Action space.}
The robot is controlled through a 13-dimensional action space
combining joint-position targets and mobile-base velocity
commands.
Each action specifies seven arm joint targets, a binary
gripper command, two head joint targets, a torso height
target, and normalized forward and angular velocity commands
for the mobile base.
Joint-position commands are executed using a PD controller.

The environment operates at \qty{10}{\hertz}, with each action
applied over two control steps at \qty{20}{\hertz}.
Policies generate action chunks and perform inference at
intervals determined by their execution horizons.

\subsection{\textbf{Memory Taxonomy}}
\label{sec:memory-taxonomy}

We categorize tasks according to the nature of the historical
information required for their successful completion.
Our taxonomy builds on concepts from psychology and
neuroscience~\citep{atkinson1968,babb2011,burgess2002,squire2004,miyake2000}
and extends the functional classification adopted in the
MIKASA-Robo family~\citep{mikasa,mikasavla}.

\begin{figure*}[t]
    \centering
    \setlength{\tabcolsep}{0pt}
    \begin{tabular}{@{}c@{\hspace{4pt}}c@{}}
        \raisebox{28pt}{
            \rotatebox{90}{\textsc{TakeItBack}}
        } &
        \includegraphics[width=0.955\textwidth]{
            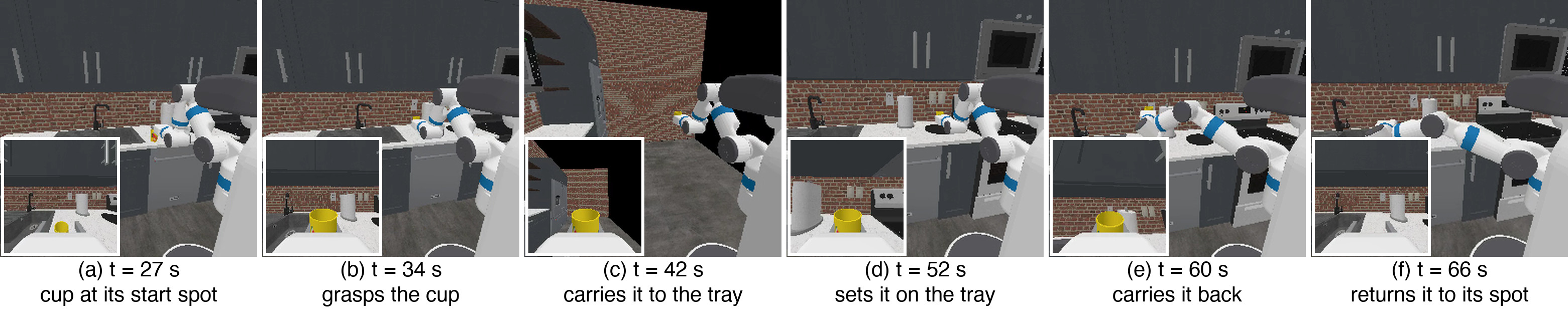
        } \\[2pt]

        \raisebox{28pt}{
            \rotatebox{90}{\textsc{DepthRecall}}
        } &
        \includegraphics[width=0.955\textwidth]{
            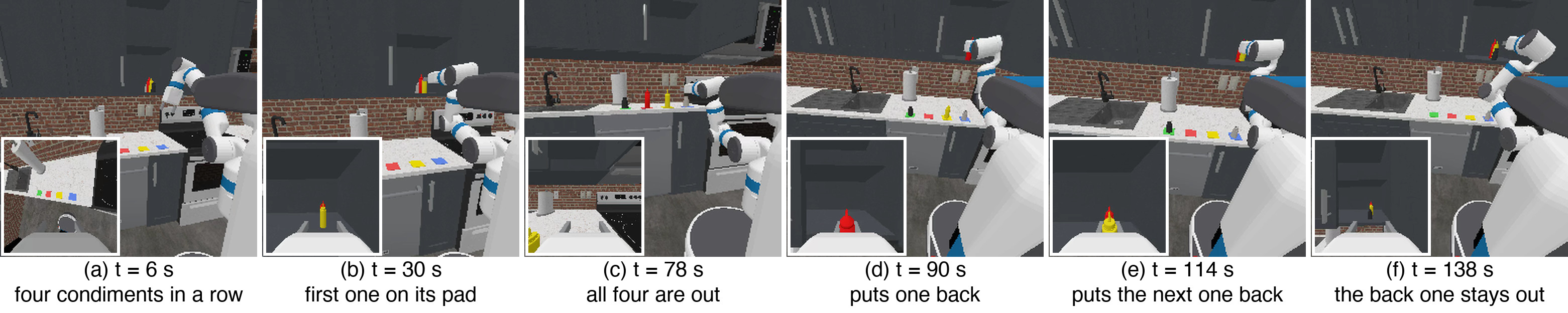
        }
    \end{tabular}

    \caption{\textbf{Memory-dependent mobile manipulation.}
    Successful demonstrations of \textsc{TakeItBack} (top)
    and \textsc{DepthRecall} (bottom).
    In \textsc{TakeItBack}, the initial cup location must be
    recalled after the cup has been transferred to the tray.
    In \textsc{DepthRecall}, removing the condiment containers
    eliminates their original spatial ordering, which must
    subsequently be reconstructed to complete the task.}
    \label{fig:takeit-depth}
\end{figure*}

\begin{itemize}
    \item \textbf{\emph{Spatial memory}} refers to retaining
    information about previously observed locations and
    spatial relations within the environment.

    \item \textbf{\emph{Object memory}} refers to retaining
    information about the identities and attributes of
    previously observed objects.

    \item \textbf{\emph{Sequential memory}} refers to retaining
    the relative order of objects, events, or actions.

    \item \textbf{\emph{Checklist memory}} refers to maintaining
    and updating information about completed interactions
    or subtasks.
\end{itemize}

While sequential memory preserves the ordering of historical
events, checklist memory primarily represents which events
have already occurred.
These categories characterize the principal memory requirements
of the tasks and are not mutually exclusive.
Individual tasks may combine multiple memory types or introduce
complementary requirements, such as retaining a numerical
target and tracking the number of executed actions.

\subsection{\textbf{Memory-Intensive Tasks}}

The benchmark comprises six tasks that instantiate different
memory requirements within household manipulation scenarios.
Table~\ref{tab:tasks} summarizes their principal memory types,
task objectives, and demonstration lengths.

\noindent\textbf{Spatial memory.}
In \textsc{TakeItBack}, the robot transfers a cup from a
randomized initial position to a tray and subsequently
returns it to its original location.
Moving the cup removes the direct visual indication of its
initial position, making spatial recall relevant to the final
placement.
In \textsc{SameDrawer}, the robot closes an initially open
drawer, transports an apple to a distant work area, and
returns to reopen the same drawer.
The intervening manipulation and navigation separate the
initial observation from the subsequent drawer selection.

\noindent\textbf{Object memory.}
In \textsc{Veggie}, a temporary image displayed on the
refrigerator identifies the vegetable that must subsequently
be placed in a pot.
\textsc{SeasonDish} extends this setting by presenting both
the identity of a target shaker and the required number of
shaking actions.
The robot must retain these instructions after the visual
cue disappears and execute the corresponding manipulation
while tracking its progress.

\noindent\textbf{Sequential memory.}
In \textsc{DepthRecall}, the robot removes four condiment
containers arranged along the depth of a shelf, retains
the backmost container, and returns the remaining objects
in their original order.
Designated temporary placement locations constrain the
use of object arrangement as an external record of the
original sequence.

\noindent\textbf{Checklist and spatial memory.}
In \textsc{CabinetSearch}, the robot searches for a target
object distributed among five closed cabinet compartments.
After inspecting an empty compartment, it must close it
and return to a designated reference location before
continuing the search.
Reopening an inspected compartment constitutes task failure,
making the accumulated history of visited compartments
relevant to subsequent search decisions.

Figure~\ref{fig:takeit-depth} illustrates the memory-dependent
interactions in \textsc{TakeItBack} and \textsc{DepthRecall}.
A complete specification of the task instructions,
randomization procedures, success criteria, and termination
conditions is provided in Appendix~\ref{app:tasks}.

\begin{table*}[t]
    \centering
    \caption{\textbf{Overview of MIKASA-Robo-MMM tasks.}
    Each task is characterized by its principal memory requirements,
    task-relevant historical information, mechanism of information
    loss, and manipulation objective. Mean demonstration lengths
    are reported in environment steps at \qty{10}{\hertz}.}
    \label{tab:tasks}

    \footnotesize
    \setlength{\tabcolsep}{5pt}
    \renewcommand{\arraystretch}{1.35}

    \begin{tabularx}{\textwidth}{
        @{}l
        l
        >{\raggedright\arraybackslash}p{0.16\textwidth}
        >{\raggedright\arraybackslash}p{0.14\textwidth}
        >{\raggedright\arraybackslash}X
        r@{}
    }
        \hline

        \textbf{Task}
        & \makecell[l]{\textbf{Memory}\\\textbf{type(s)}}
        & \makecell[l]{\textbf{Information}\\\textbf{to retain}}
        & \makecell[l]{\textbf{Information}\\\textbf{loss}}
        & \textbf{Task description}
        & \makecell[r]{\textbf{Mean}\\\textbf{steps}}
        \\

        \hline

        \textsc{TakeItBack}
        & Spatial
        & Initial cup position
        & Object relocation
        & Transfer a cup to a tray and return it
          to its original position.
        & 554
        \\

        \textsc{SameDrawer}
        & Spatial
        & Initially open drawer
        & Drawer closure
        & Close the drawer, transfer an apple to a plate,
          and reopen the same drawer.
        & 560
        \\

        \textsc{CabinetSearch}
        & \makecell[l]{Spatial\\Checklist}
        & Previously inspected cabinets
        & Compartment closure and navigation
        & Find a cola can among five cabinets without
          reopening an inspected cabinet.
        & 1{,}314
        \\

        \textsc{Veggie}
        & Object
        & Target vegetable identity
        & Temporary cue removal
        & Place the vegetable indicated by an earlier
          visual cue into a pot.
        & 930
        \\

        \textsc{SeasonDish}
        & Object
        & Target shaker identity and action count
        & Temporary cue removal
        & Shake the indicated shaker the specified
          number of times over a bowl.
        & 574
        \\

        \textsc{DepthRecall}
        & Sequential
        & Original front-to-back ordering of condiments
        & Object rearrangement
        & Remove four condiments from a shelf,
          retain the backmost one, and restore
          the remaining objects.
        & 1{,}377
        \\

        \hline

    \end{tabularx}
\end{table*}

\section{Dataset}
\label{sec:dataset}

To facilitate supervised fine-tuning and evaluation of
memory-augmented VLA policies, we provide 1{,}000 successful
demonstrations for each of the six benchmark tasks.
The resulting dataset comprises 6{,}000 trajectories and
5{,}308{,}563 environment steps, corresponding to approximately
147 hours of robot interaction recorded at \qty{10}{\hertz}.

\subsection{\textbf{Demonstration Generation}}

We develop task-specific scripted motion planners using
MPlib, the motion planning library integrated into
ManiSkill3~\citep{maniskill3}.
Each planner implements a sequence of navigation and
manipulation primitives conditioned on the initial scene
configuration and the task objective.

To increase trajectory diversity, demonstrations are collected
under randomized environment configurations, with additional
perturbations applied to motion-planning waypoints and action
execution.
The planners achieve an approximate success rate of 85\%
across the benchmark.
Only trajectories satisfying the corresponding task-specific
success criteria are retained in the dataset.

\subsection{\textbf{Trajectory Processing}}

To improve data collection efficiency, trajectories are
initially generated using state-based simulation.
Successful executions are subsequently replayed in RGB
observation mode to obtain visual and proprioceptive
observations aligned with the recorded actions.

The original trajectories are collected at a control
frequency of \qty{20}{\hertz} and subsequently downsampled
to \qty{10}{\hertz}.
Consecutive control commands are aggregated through
temporal averaging to preserve the commanded motion
over each downsampled interval, particularly during
mobile-base navigation.

\subsection{\textbf{Dataset Organization and Release}}

The dataset is publicly available on Hugging Face in
LeRobotDataset v3.0 format~\citep{lerobot}.\footnote{\url{https://huggingface.co/datasets/MIKASA-Robo-MMM/MIKASA-Robo-MMM}}
Demonstrations from all six tasks are consolidated into
a unified dataset containing synchronized RGB observations,
proprioceptive measurements, actions, and task-specific
language instructions.

For reproducible evaluation, we additionally provide
100 held-out, planner-verified environment seeds per task,
disjoint from the training seeds.

The dataset also includes the global pose of the mobile
base as auxiliary diagnostic information in the
\texttt{global\_state} field.
This information is reserved for trajectory analysis
and excluded from policy inputs in the benchmark
evaluation protocol.

\section{Experimental Evaluation}
\label{sec:baselines}

We evaluate six VLA policies to investigate the effectiveness
of existing memory mechanisms in mobile manipulation.
The comparison includes a memoryless baseline and five
memory-augmented architectures that differ in how they
represent, update, and retrieve historical information.
All policies are fine-tuned on the provided demonstrations
and evaluated through closed-loop interaction with the
benchmark environments.

\subsection{\textbf{Baseline Policies}}

\noindent\textbf{Memoryless $\pi_{0.5}$.}
We use $\pi_{0.5}$~\citep{pi05} without an explicit memory
mechanism as a reactive baseline, conditioning action
generation on the current observation and language
instruction.

\noindent\textbf{FrameSamp+Modul and TokenDrop+Modul.}
These memory-augmented variants of $\pi_{0.5}$,
introduced in RoboMME~\citep{robomme}, retain perceptual
information from previous observations through frame
sampling and visual token selection, respectively.
For TokenDrop+Modul, historical features are extracted
from the left head-mounted camera, following the
single-camera memory configuration of the original
architecture, while all three current camera observations
remain available to the policy.

\noindent\textbf{HAMLET.}
HAMLET~\citep{hamlet} maintains a compressed representation
of temporal information through learned memory tokens.
We employ GR00T N1.6 as its backbone instead of the
original GR00T N1.5~\citep{gr00t} to accommodate
the heterogeneous camera resolutions of DS-Fetch.

\noindent\textbf{MemER.}
MemER~\citep{memer} follows a hierarchical architecture
in which historical keyframes inform subgoal prediction
by a high-level vision-language model, while a low-level
VLA executes the resulting subgoals.
We use Qwen3-VL-4B~\citep{qwen3vl} for high-level
reasoning and adopt the low-level policy configuration
from the RoboMME reproduction.

\noindent\textbf{MemoryVLA.}
MemoryVLA~\citep{memoryvla} maintains a latent memory bank
that supports the retrieval and consolidation of historical
features, providing the policy with access to information
accumulated throughout the episode.

\subsection{\textbf{Fine-Tuning Protocol}}

All architectures are adapted to the observation and
action spaces of DS-Fetch, including the three-camera
configuration, 12-dimensional proprioceptive observations,
and 13-dimensional action representation at
\qty{10}{\hertz}.

We follow the training configurations reported in the
original publications and their available implementations,
preserving the respective memory architectures while
adapting their embodiment-specific components to our
benchmark.
Each policy is fine-tuned using four GPUs.
Detailed training configurations, optimization parameters,
and architecture-specific adaptations are provided in
Appendix~\ref{app:hyperparams}.

\subsection{\textbf{Evaluation Protocol}}

Each policy is evaluated on 100 held-out environment
seeds per task, shared across all evaluated architectures.
The evaluation seeds are disjoint from the corresponding
training seeds and are verified through successful
execution by the task-specific scripted planners.

For each task, we report the success rate as the empirical
estimate of the objective introduced in
Section~\ref{sec:background}:
\[
    \widehat{J}(\pi)
    =
    \frac{1}{N}
    \sum_{i=1}^{N} S(\tau_i),
    \qquad N=100,
\]
where $S(\tau_i)$ indicates successful completion of
the $i$-th evaluation episode.
Overall performance is reported as the unweighted mean
success rate across all benchmark tasks.

The evaluation uses task-specific episode horizons and
termination conditions described in
Appendix~\ref{app:tasks}.
All evaluation seeds are distributed with the benchmark
to facilitate reproducible comparisons.

\subsection{\textbf{Experimental Results}}

Table~\ref{tab:results} presents the performance of the
evaluated policies across the six benchmark tasks.
HAMLET achieves the highest mean success rate of 10.7\%,
followed by MemER at 7.2\%, compared with 1.8\%
for the memoryless $\pi_{0.5}$ baseline.

The largest improvements are observed in
\textsc{TakeItBack}, where MemER and HAMLET achieve
success rates of 34\% and 30\%, respectively,
compared with 7\% for the memoryless baseline.
HAMLET additionally achieves the highest performance
on \textsc{CabinetSearch} and \textsc{Veggie},
indicating its ability to complete tasks involving
both accumulated interaction history and temporary
visual cues.

FrameSamp+Modul, TokenDrop+Modul, and MemoryVLA achieve
substantially lower aggregate performance, demonstrating
that the effectiveness of existing memory-augmented
architectures varies considerably across the benchmark.

Despite the improvements achieved by HAMLET and MemER,
\textsc{SameDrawer}, \textsc{SeasonDish}, and
\textsc{DepthRecall} remain unsolved by all evaluated
policies.
These results highlight persistent challenges in
coordinating memory-dependent decisions with extended
sequences of navigation and manipulation.

\begin{table}[t]
    \centering
    \caption{\textbf{Benchmark results.}
    Success rates (\%) over 100 evaluation episodes per
    task. The final column reports the unweighted mean
    across all six tasks.}
    \label{tab:results}
    \setlength{\tabcolsep}{2pt}
    \renewcommand{\arraystretch}{1.15}

    \resizebox{\columnwidth}{!}{%
    \begin{tabular}{@{}lccccccc@{}}
        \hline
        & \textbf{TakeIt} & \textbf{Same} & \textbf{Cabinet} & &
        \textbf{Season} & \textbf{Depth} & \\
        \textbf{Model}
        & \textbf{Back} & \textbf{Drawer} & \textbf{Search} & \textbf{Veggie}
        & \textbf{Dish} & \textbf{Recall} & \textbf{Mean} \\
        \hline
        $\pi_{0.5}$
        & 7 & 0 & 4 & 0 & 0 & 0 & 1.8 \\

        FrameSamp+Modul
        & 16 & 0 & 3 & 0 & 0 & 0 & 3.2 \\

        TokenDrop+Modul
        & 8 & 0 & 3 & 0 & 0 & 0 & 1.8 \\

        HAMLET
        & 30 & 0 & \textbf{21} & \textbf{13}
        & 0 & 0 & \textbf{10.7} \\

        MemER
        & \textbf{34} & 0 & 6 & 3 & 0 & 0 & 7.2 \\

        MemoryVLA
        & 3 & 0 & 0 & 0 & 0 & 0 & 0.5 \\
        \hline
    \end{tabular}}
\end{table}

\subsection{\textbf{Failure Analysis}}
\label{sec:failure}

Qualitative analysis of evaluation trajectories reveals
two recurring failure modes associated with manipulation
execution and the utilization of historical information.

\noindent\textbf{Manipulation execution.}
The coordination of mobile-base motion, arm control,
and interaction with articulated objects remains
challenging for existing VLA policies.
Failures frequently occur during object acquisition,
placement, and cabinet opening.
HAMLET exhibits comparatively stronger manipulation
behavior, particularly during grasping and interaction
with articulated fixtures, contributing to its higher
overall task completion rate.

\noindent\textbf{Utilization of historical information.}
Several policies exhibit behavior dominated by the
current observation even when earlier observations
contain information relevant to subsequent decisions.
In \textsc{Veggie}, policies frequently select objects
based on their current visual accessibility rather
than the previously presented target cue.
Similarly, in \textsc{SameDrawer}, HAMLET predominantly
selects the bottom drawer regardless of which drawer
was initially open.

These behavioral patterns highlight the importance of
integrating historical information into action generation.
Effective memory-augmented mobile manipulation requires
both the retention of task-relevant information and
its consistent utilization across changing viewpoints,
intermediate interactions, and extended action sequences.

\section{Conclusion}
\label{sec:conclusion}

We introduced MIKASA-Robo-MMM, a benchmark for evaluating
memory in mobile manipulation, comprising six memory-intensive
household tasks and 6{,}000 planner-generated demonstrations.
By embedding history-dependent decisions within navigation
and manipulation, the benchmark extends the study of memory
in robotic policies beyond fixed-base settings.
Our evaluation of five memory-augmented VLA architectures
and a memoryless baseline reveals substantial challenges:
the best-performing policy achieves a mean success rate
of 10.7\%, compared with 1.8\% for the memoryless baseline,
while three tasks remain unsolved by all evaluated policies.
These results motivate the development of memory architectures
that not only retain task-relevant historical information
but also effectively integrate it into action generation
across changing viewpoints and extended sequences of
physical interactions.

\section*{Acknowledgments}
\bibliographystyle{plainnat}
\bibliography{references}

\clearpage
\onecolumn
\appendices
\newcommand{\apptodo}[1]{\par\smallskip\noindent\textcolor{red}{\textbf{TODO (authors).} #1}\par\smallskip}

\section*{Appendix Contents}
{\small
\newcommand{\appendixentry}[3]{%
  \noindent\hspace*{#1}%
  \hyperref[#2]{\ref*{#2}\enspace #3\dotfill\pageref*{#2}}\par}
\appendixentry{0em}{app:tasks}{Task Specifications and Evaluation Conditions}
\appendixentry{2em}{app:common-protocol}{Common Protocol}
\appendixentry{2em}{app:takeitback}{TakeItBack}
\appendixentry{2em}{app:samedrawer}{SameDrawer}
\appendixentry{2em}{app:cabinetsearch}{CabinetSearch}
\appendixentry{2em}{app:veggie}{Veggie}
\appendixentry{2em}{app:seasondish}{SeasonDish}
\appendixentry{2em}{app:depthrecall}{DepthRecall}
\appendixentry{2em}{app:demonstrations}{Demonstration Collection Statistics}
\smallskip
\appendixentry{0em}{app:hyperparams}{Baseline Configurations and Training Hyperparameters}
\appendixentry{2em}{app:pi05}{Memoryless $\pi_{0.5}$}
\appendixentry{2em}{app:memer}{MemER}
\appendixentry{2em}{app:framesamp-tokendrop}{FrameSamp+Modul and TokenDrop+Modul}
\appendixentry{2em}{app:hamlet}{HAMLET}
\appendixentry{2em}{app:memoryvla}{MemoryVLA}
}
\medskip

\makeatletter
\newenvironment{taskfigure}
  {\par\medskip\noindent\minipage{\linewidth}\def\@captype{figure}}%
  {\endminipage\par\medskip}
\newcommand{\rolloutrow}[1]{%
  \includegraphics[width=0.4984\linewidth,trim=0 626bp 0 0,clip]{#1}%
  \hfill
  \includegraphics[width=0.4984\linewidth,trim=0 2bp 0 624bp,clip]{#1}\par
}
\newenvironment{appconfig}[2]
  {\par\medskip\noindent\minipage{\linewidth}\def\@captype{table}%
   \caption{#1}\label{#2}\centering\small
   \setlength{\tabcolsep}{6pt}\renewcommand{\arraystretch}{1.15}%
   \begin{tabular}{@{}>{\raggedright\arraybackslash}p{0.27\linewidth}%
     >{\raggedright\arraybackslash}p{\dimexpr0.73\linewidth-2\tabcolsep\relax}@{}}%
   \hline\textbf{Parameter} & \textbf{Configuration} \\\hline}%
  {\\\hline\end{tabular}\endminipage\par\medskip}
\makeatother

\section{Task Specifications and Evaluation Conditions}
\label{app:tasks}

This section describes the task-specific scene configurations, memory
requirements, randomization procedures, success criteria, and scripted
planners of MIKASA-Robo-MMM. All tasks are implemented in the small
one-wall RoboCasa kitchen with the industrial style and share the DS-Fetch
interface described in Section~\ref{sec:obs-actions}. Demonstrations are
collected with assisted grasping as described in
Section~\ref{sec:design-choices}.

\subsection{\textbf{Common Protocol}}
\label{app:common-protocol}

\taskfield{Demonstration generation} Task-specific scripted planners use
simulator state information to generate navigation and manipulation
trajectories. Successful executions are replayed in RGB observation mode
and processed at \qty{10}{\hertz}, as described in Section~\ref{sec:dataset}.
The privileged state used by the planners is excluded from policy inputs.
The dataset provides 1{,}000 successful demonstrations per task.

\taskfield{Evaluation instances and horizons} Each task provides 100
held-out seeds in \texttt{validation\_seeds.json}, disjoint from its
training seeds. Each evaluation seed was executed successfully by its
scripted planner before inclusion. An instance is identified by its
(task, seed) pair. The evaluation horizon is twice the maximum length of
the released demonstrations for the task and overrides the registered
environment horizon. Unless otherwise stated, lengths are reported in
environment steps at \qty{10}{\hertz}. The underlying control frequency is
\qty{20}{\hertz}. Task-specific irreversible violations and termination
conditions are distinguished below.

\taskfield{Temporal statistics} Intervals between task events are
estimated using subtask annotations of scripted demonstrations, with a
resolution of 1.6\,s. We specify the events and the number of trajectories
used for each reported statistic. These intervals describe the timing of
scripted executions rather than fixed deadlines imposed on the policy.

\taskfield{Random-choice references} For selected tasks, we report the
probability of a correct decision under a specified uniform-choice
strategy, assuming successful execution of the required motor actions.
These probabilities characterize the stated random-choice strategies,
rather than the performance of an optimal reactive policy.

\begin{table}[!htbp]
  \centering
  \caption{\textbf{Demonstration statistics and evaluation horizons.}
    Lengths and evaluation horizons are expressed in environment steps at
  \qty{10}{\hertz}.}
  \label{tab:app-task-stats}
  \small
  \setlength{\tabcolsep}{5pt}
  \renewcommand{\arraystretch}{1.15}
  \begin{tabular}{@{}lrrrrr@{}}
    \hline
    \textbf{Task} & \textbf{Demos} & \textbf{Min.} & \textbf{Mean} &
    \textbf{Max.} & \textbf{Eval. horizon} \\
    \hline
    \textsc{TakeItBack} & 1{,}000 & 391 & 554 & 835 & 1{,}670 \\
    \textsc{SameDrawer} & 1{,}000 & 520 & 560 & 690 & 1{,}380 \\
    \textsc{CabinetSearch} & 1{,}000 & 404 & 1{,}314 & 2{,}371 & 4{,}742 \\
    \textsc{Veggie} & 1{,}000 & 883 & 930 & 1{,}089 & 2{,}178 \\
    \textsc{SeasonDish} & 1{,}000 & 337 & 574 & 828 & 1{,}656 \\
    \textsc{DepthRecall} & 1{,}000 & 1{,}343 & 1{,}377 & 1{,}493 & 2{,}986 \\
    \hline
  \end{tabular}
\end{table}
\subsection{\textbf{TakeItBack}}
\label{app:takeitback}

\begin{taskfigure}
  \centering
  \rolloutrow{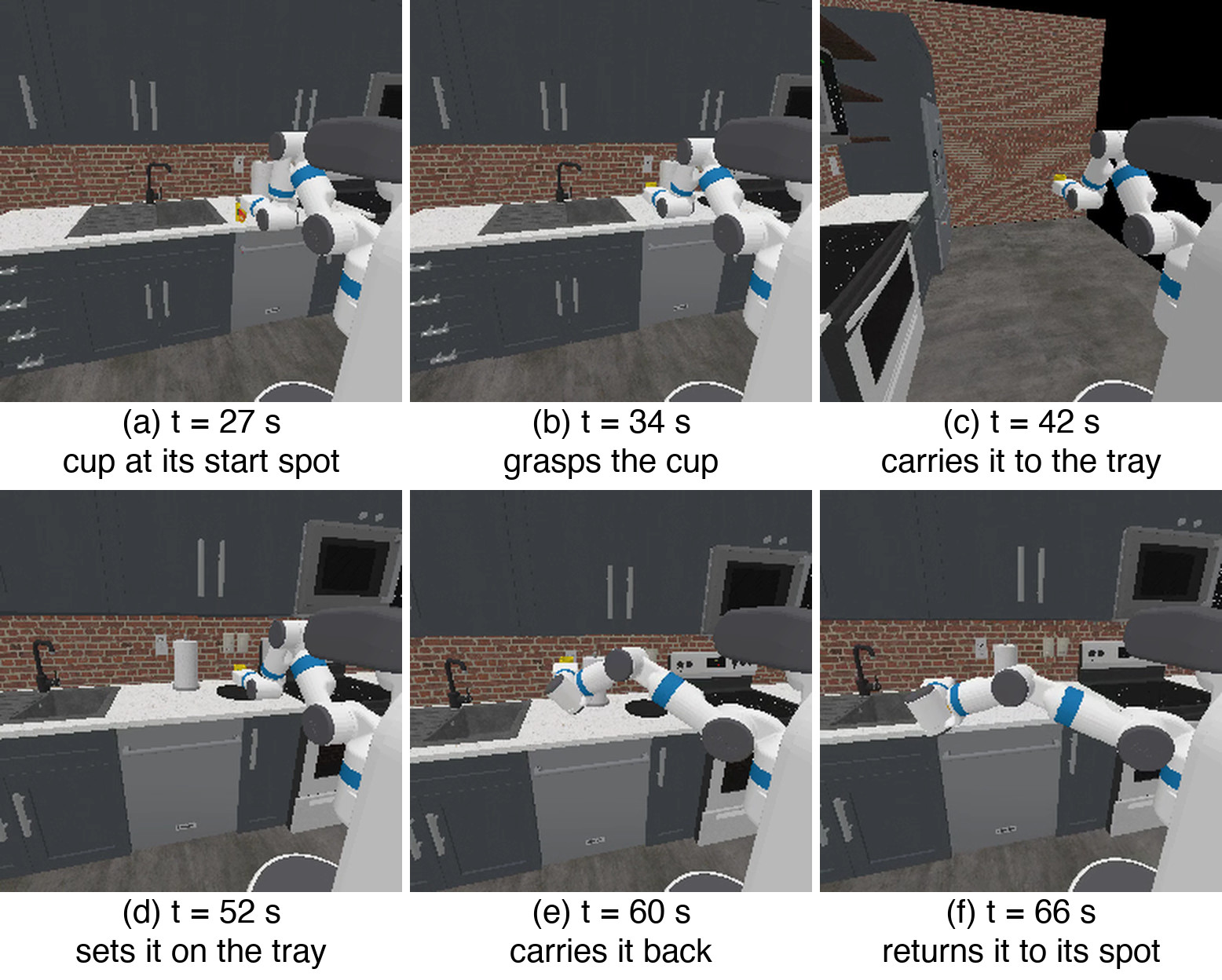}
  \caption{\textbf{TakeItBack.} Example of transferring a cup to a tray and
    restoring its original position on the counter. The initial cup location
    is the spatial information relevant to the final placement. Demonstration
  985, left head-mounted camera.}
  \label{fig:app-takeitback}
\end{taskfigure}

\taskfield{Scene and objective} A cup is initialized on the front strip of
the counter, with a 0.28\,m baking tray between the strip and the wall.
The robot transfers the cup to the tray and subsequently restores its
original position.

\taskfield{Language instruction} ``Move the cup from the counter onto the
tray, then return it to its original position on the counter.''

\taskfield{Memory requirements} The initial cup position provides the
spatial reference for the final placement. Once the cup is removed, its
original location has no explicit visual marker. In a sample of 20 released
demonstrations, the return phase begins 11--24\,s (median 13\,s) after the
start of transport to the tray.

\taskfield{Initial-state randomization} The cup position is sampled uniformly
from 0.68\,m of available counter space, distributed over three segments
spanning 0.85\,m. The base position is sampled uniformly from 168 free cells
of a 0.10\,m floor grid spanning $2.0\times0.7$\,m along and across the
counter. Its heading is sampled uniformly from admissible orientations in
$2^\circ$ increments, subject to free floor 1.13\,m ahead. The initial torso
and arm joint positions deviate from their rest configurations by at most
3\,cm and 0.03\,rad, respectively. The tray position varies by at most 7\,cm.

\taskfield{Success and termination} Success requires two ordered conditions:
(i) the cup has previously rested on the tray while released, and (ii) it
currently rests released within 0.15\,m horizontally and 0.10\,m vertically
of its initial position. The tray region is defined by its footprint
expanded by 5\,cm and a vertical tolerance of 0.10\,m. The episode terminates
at the first successful step. The robot can recover from a dropped cup
within the evaluation horizon of 1{,}670 environment steps (167.0\,s).

\taskfield{Scripted planner}
\emph{Approach and acquisition.} Raise the arm to a ready configuration.
If the cup is farther than 0.9\,m, align the base with the counter and drive
until the cup is abeam. Face the cup and stop 0.25\,m from the counter.
Lower the torso, close the gripper around the cup, and raise the torso by
0.17\,m.
\emph{Transfer.} Move the cup to a position 0.15\,m above the tray, typically
through coordinated arm motion and forward base motion, or by driving.
Lower the torso and release the cup onto the tray.
\emph{Restoration.} Regrasp the cup, lift it by 0.17\,m, and transport it to
0.17\,m above the stored initial position. Lower and release the cup onto
the counter. A failed planner check aborts the attempt without a retry.

\subsection{\textbf{SameDrawer}}
\label{app:samedrawer}

\begin{taskfigure}
  \centering
  \rolloutrow{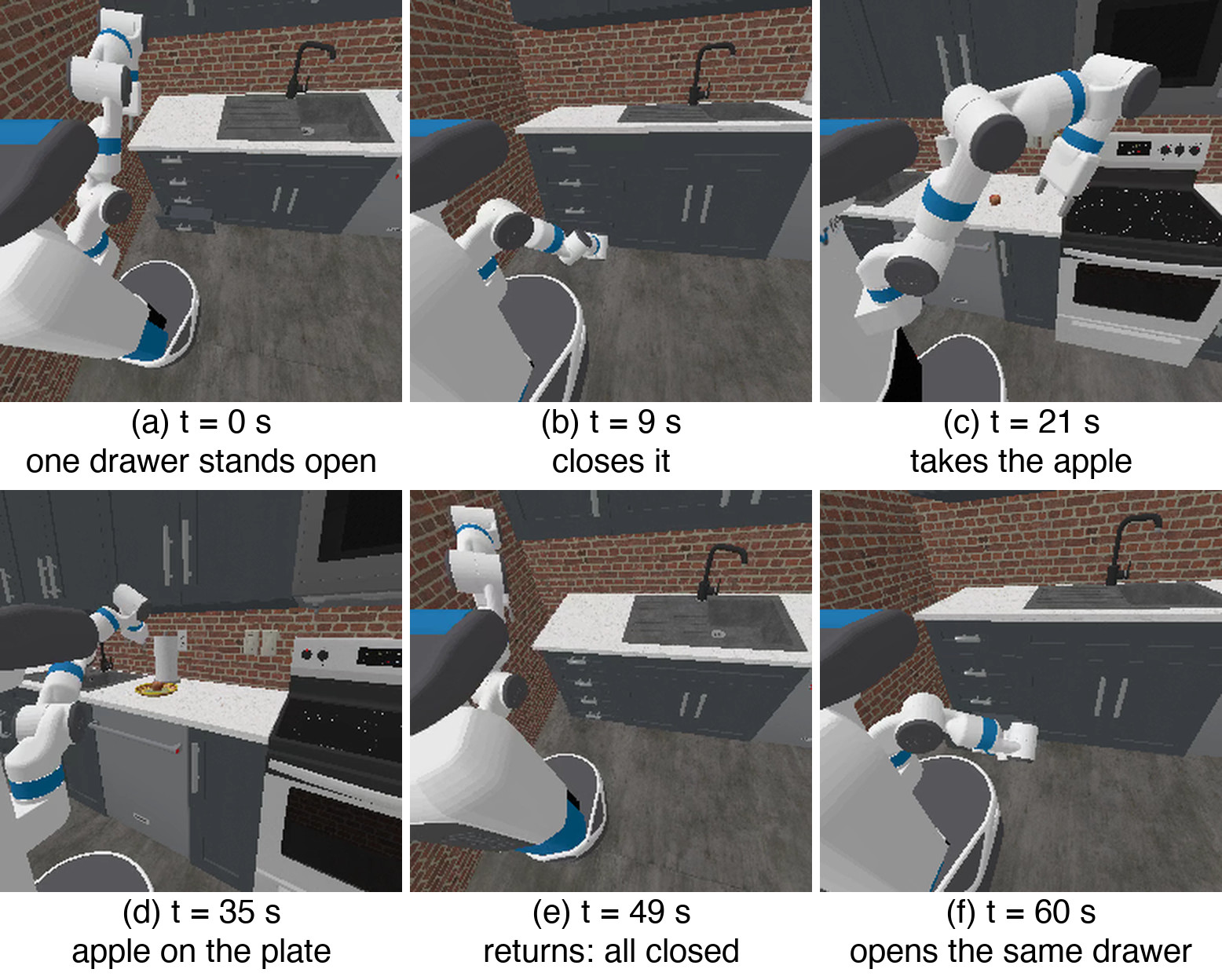}
  \caption{\textbf{SameDrawer.} The robot closes the initially open drawer,
    performs an intermediate object-placement task, and returns to reopen
    the selected drawer. First released demonstration, right head-mounted
  camera.}
  \label{fig:app-samedrawer}
\end{taskfigure}

\taskfield{Scene and objective} Four vertically stacked drawers are located
at one end of the counter, exactly one of which is initially open. An apple
and a plate are located farther along the counter. The robot must close
the open drawer, place the apple on the plate, and reopen that drawer.

\taskfield{Language instruction} ``Remember which drawer is open and close
it. Keep it closed while putting the apple on the plate at the far counter,
then return and open the same drawer. Keep the apple on the plate until the
end and leave every other drawer closed.''

\taskfield{Memory requirements} The initially open drawer determines the
final drawer selection. To remove an explicit mechanical indicator of prior
interaction, a drawer whose opening is less than 5\,mm and which is not being
pulled open is set to its fully closed joint position. After closing the
initial drawer, the robot travels approximately 1.8\,m to the apple and
returns after placement. In 100 scripted evaluation episodes, reopening
begins 34--42\,s (median 35\,s) after drawer closure.

\taskfield{Initial-state randomization} The initially open drawer is sampled
uniformly from the four drawers, with an opening between 0.10 and 0.16\,m.
The apple, plate, and base positions vary by up to 2\,cm per horizontal
axis, and the base heading by up to 0.03\,rad. Initial deviations from the
rest configuration are bounded by 0.04\,rad for the arm, 1\,cm for the torso,
and 0.05\,rad for the head joints.

\taskfield{Success and termination} Success requires closing the initially
open drawer, releasing the apple onto the plate, and subsequently reopening
the same drawer by at least 0.10\,m, in that order. The resulting configuration,
with the apple on the plate and all other drawers closed, must persist for
0.5\,s. The following violations irreversibly invalidate the episode:
\begin{itemize}
  \item Opening a different drawer by more than 2\,cm.
  \item Placing the apple before closing the initial drawer, or reopening
    the initial drawer before placing the apple.
  \item Removing or regrasping the apple after its release onto the plate.
\end{itemize}
Invalidated episodes continue until the evaluation horizon of 1{,}380
environment steps (138.0\,s).

\taskfield{Random-choice reference} Under a single uniform choice among the
four drawers, the probability of selecting the correct drawer is $1/4$.

\taskfield{Scripted planner}
\emph{Drawer closure.} Orient the head toward the drawer column and approach
it when required for a lower drawer. Push the selected drawer along its
slider with an open hand until its opening is within 5\,mm, allowing one
re-aiming attempt after loss of palm contact. Retract the hand by 0.24\,m and
fold the arm.
\emph{Intermediate manipulation.} Drive to the apple, approach from behind
and above at $45^\circ$, grasp it, and lift it by 7\,cm. Transfer the apple
over the plate, lower it, and release it after the arm settles.
\emph{Drawer reopening.} Fold the arm, return to the drawer column, grasp
the stored target drawer handle, and pull it 2--8\,cm beyond the 0.10\,m
opening threshold. Release the handle and hold the configuration until
success.

\subsection{\textbf{CabinetSearch}}
\label{app:cabinetsearch}

\begin{taskfigure}
  \centering
  \rolloutrow{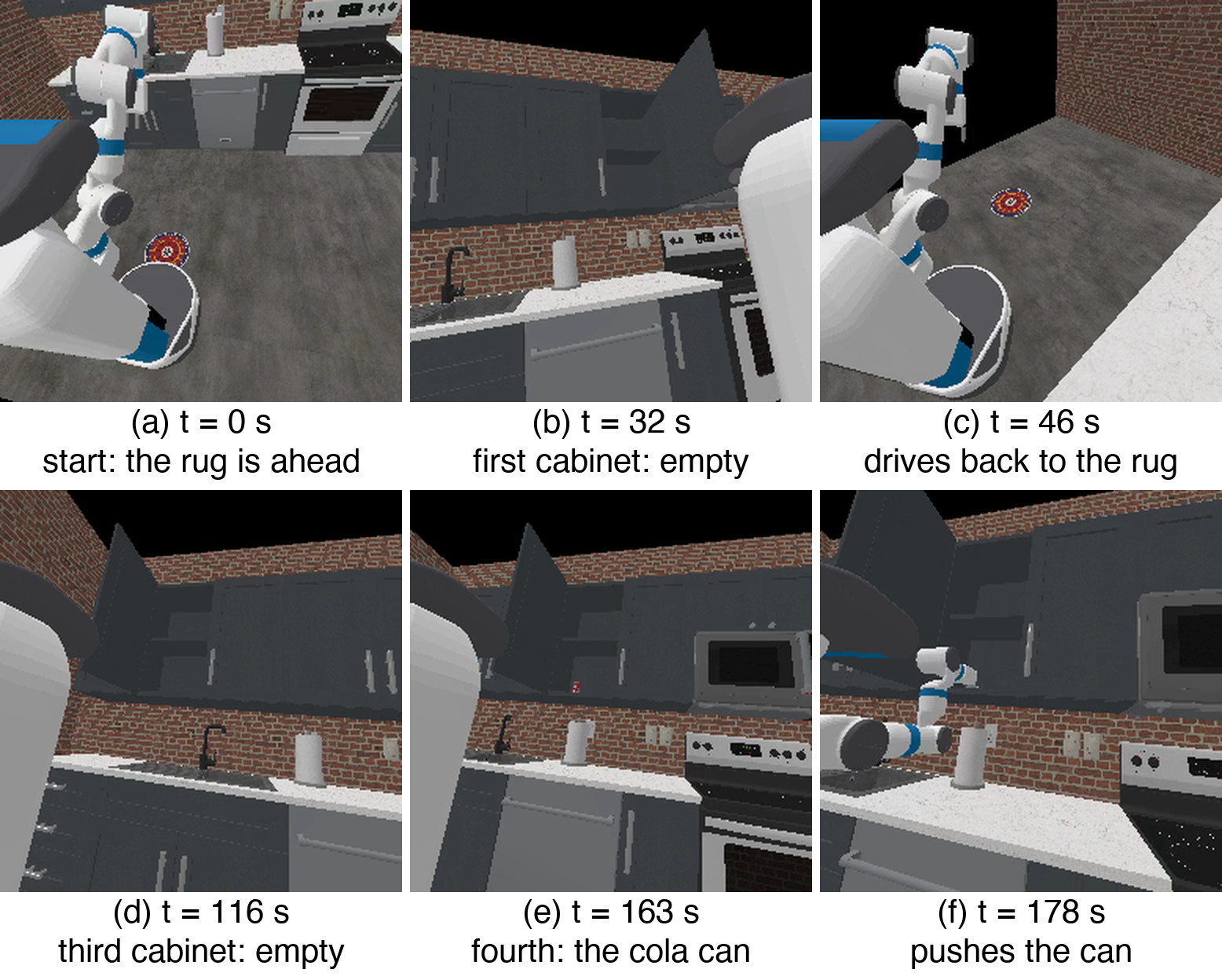}
  \caption{\textbf{CabinetSearch.} Iterative search over closed cabinet
    compartments, with a return to the rug between unsuccessful inspections.
    The target is located in the fourth inspected compartment in this example.
    In this first released demonstration, frame (b) uses the left
    head-mounted camera
  and the remaining frames use the right head-mounted camera.}
  \label{fig:app-cabinetsearch}
\end{taskfigure}

\taskfield{Scene and objective} A cola can is placed in one of five closed
wall-cabinet compartments, each with its own door: one single-door cabinet
and two partitioned double-door cabinets. A rug of radius 0.20\,m provides
a reference location in front of the cabinets. The robot is initialized
0.5\,m behind the nominal rug position. The task requires locating and
pushing the can without repeated inspections.

\taskfield{Language instruction} ``Find the cola can in the wall cabinets
and gently push it. Start at the rug. Close empty cabinets and return
directly to the rug before searching another. Never inspect a cabinet twice.
Finish after pushing the can.'' The three instruction variants used during
data collection are standardized to this phrasing in the released dataset.

\taskfield{Memory requirements} The target is visible only when its
compartment is open. After inspecting an empty compartment, the robot
closes its door and returns to the rug before selecting another compartment.
The sequence therefore requires maintaining a record of previous inspections
when the search extends beyond the first compartment. Closed doors carry
no explicit inspection markers and retain their physical hinge positions
within the closure tolerance of 0.01\,rad. Returning to the rug constrains
spatial cues that could otherwise encode progress through a fixed search
order.

In 100 scripted evaluation episodes, successive compartment openings begin
13--18\,s (median 14\,s) after closing the preceding compartment. The
interval measured from the first closure reaches 154\,s (median 59\,s).
An episode in which the target is found during the first inspection does
not require maintaining an inspection history.

\taskfield{Initial-state randomization} The target compartment is sampled
uniformly among the five. The can position varies by up to 6\,cm laterally
and 2\,cm in depth. The rug position varies by up to 0.25\,m along the
cabinets and 0.08\,m perpendicular to them. The base position varies by up
to 8\,cm per horizontal axis and its yaw by up to 0.10\,rad. Initial arm,
torso, and head deviations from rest are bounded by 0.04\,rad, 1\,cm, and
0.05\,rad, respectively.

\taskfield{Success and termination} Success requires opening the target
compartment by at least 1.2\,rad and displacing the can by at least 2\,cm,
without a preceding violation. Opening a previously inspected compartment,
opening one while another is open, or opening one without first stopping on
the rug with a heading within $35^\circ$ of the cabinets terminates the
episode in failure. An additional pull of the most recently opened door is
permitted before it reaches 1.2\,rad. A door is classified as open when its
angle exceeds 0.05\,rad, and as closed when it remains within 0.01\,rad of
its closed position for 0.5\,s with the hand at least 0.25\,m from its
handle. The evaluation horizon is 4{,}742 environment steps (474.2\,s).

\taskfield{Scripted planner}
\emph{Search initialization.} Fold the arm, navigate to the rug, and select
the next compartment according to a seeded permutation.
\emph{Inspection.} Approach the handle and grasp it, allowing one retry
1.5\,cm deeper if necessary. Open the door toward 1.75\,rad with coordinated
base motion, release the handle, and inspect the compartment for 0.5\,s.
\emph{Search continuation.} For an empty compartment, close the door with
a fist, reducing speed over the final 0.10\,rad. Withdraw, fold the arm,
return to the rug, and proceed to the next compartment.
\emph{Target interaction.} Upon locating the can, approach it, raise the
torso, and push it with a fist, allowing up to three attempts. The single
door uses a target opening of 1.28\,rad and is closed using the same grip.

\taskfield{Search-depth distribution} Among the 1{,}000 released demonstrations,
207, 189, 193, 217, and 194 episodes inspect one, two, three, four, and five
compartments, respectively.

\subsection{\textbf{Veggie}}
\label{app:veggie}

\begin{taskfigure}
  \centering
  \rolloutrow{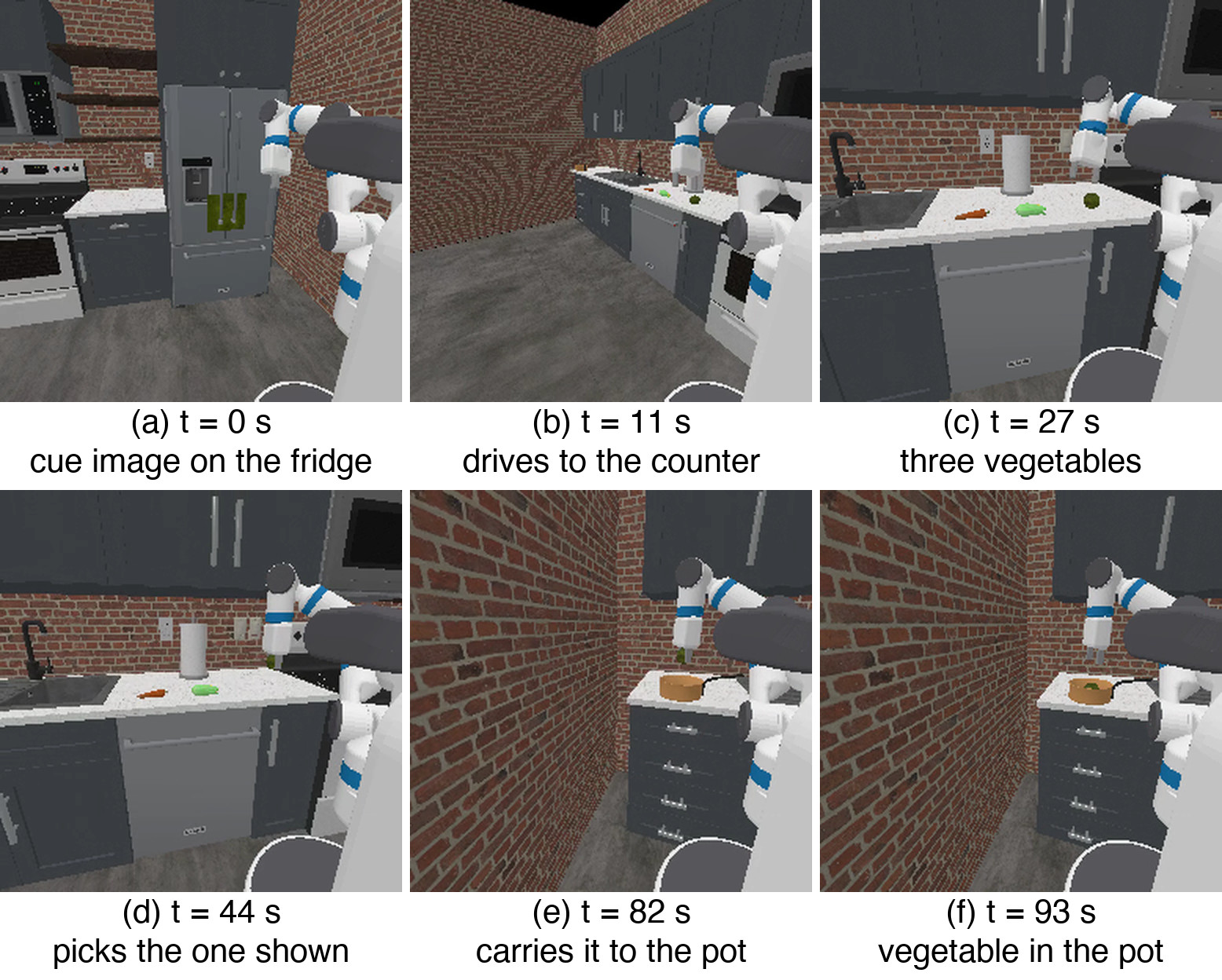}
  \caption{\textbf{Veggie.} A temporary image on the refrigerator specifies
    the target vegetable for subsequent selection and transport to the pot.
  First released demonstration, left head-mounted camera.}
  \label{fig:app-veggie}
\end{taskfigure}

\taskfield{Scene and objective} Three vegetables of different kinds are
placed on the counter, with a pot at its far end. A square image displayed
on the refrigerator shows the texture of the target vegetable's 3D model.
The robot starts 1.15\,m from the refrigerator, facing the image, and must
transport the indicated vegetable to the pot.

\taskfield{Language instruction} ``Look at the picture on the refrigerator,
remember the vegetable shown, and place the matching vegetable on the plate.''
The destination referred to as the plate in the recorded instruction is
implemented as a pot in the environment. This is the exact text of the
released dataset. We did not correct it: the policies are trained on this
text, and the evaluation passes the same text as the prompt, so training and
evaluation use identical wording.

\taskfield{Memory requirements} The target image is displayed during the
first 5\,s and is subsequently removed by the environment. Navigation from
the refrigerator to the counter separates target acquisition from object
selection and grasping. In the illustrated demonstration, the left camera
loses sight of the image within 1\,s. In 20 released demonstrations, the
robot initiates navigation during cue presentation and grasps the target
30--43\,s (median 33\,s) after cue removal.

\taskfield{Initial-state randomization} Vegetable categories and models,
and the pot model, are sampled per episode. The target vegetable is sampled
uniformly among the three candidates independently of their placement.
The three counter locations, pot location, and initial robot pose remain
fixed, while the assignment of vegetables to the counter locations varies.

\taskfield{Success and termination} Success is registered at the first
step at which the target vegetable rests released within 0.15\,m horizontally
of the pot origin, with a vertical offset between $-0.06$ and $0.10$\,m.
The pot origin is defined as the center of the bounding box encompassing
the pot and its handle. No separate irreversible failure condition is
specified. The evaluation horizon is 2{,}178 environment steps (217.8\,s).

\taskfield{Random-choice reference} A single uniform choice among the three
vegetables identifies the target with probability $1/3$.

\taskfield{Scripted planner}
\emph{Approach.} Orient the base toward the target and drive to a position
0.5\,m in front of it. Lower the torso by 0.22\,m and orient the robot toward
the counter.
\emph{Acquisition.} Position the hand 0.10\,m above the vegetable with a
top-down orientation across its narrow axis. Lower the torso until the
fingertips are 5\,mm above the counter, close the gripper, and lift, verifying
that the vegetable rises by at least 3\,cm.
\emph{Placement.} Retreat 0.20\,m and drive to 0.5\,m from the pot. Raise
the torso by 0.04\,m, face the pot, and move the vegetable to 5\,cm above
the rim. Lower it to approximately 1\,cm above the pot bottom, release,
and raise the torso. A failed check aborts the attempt without retrying.

\subsection{\textbf{SeasonDish}}
\label{app:seasondish}

\begin{taskfigure}
  \centering
  \rolloutrow{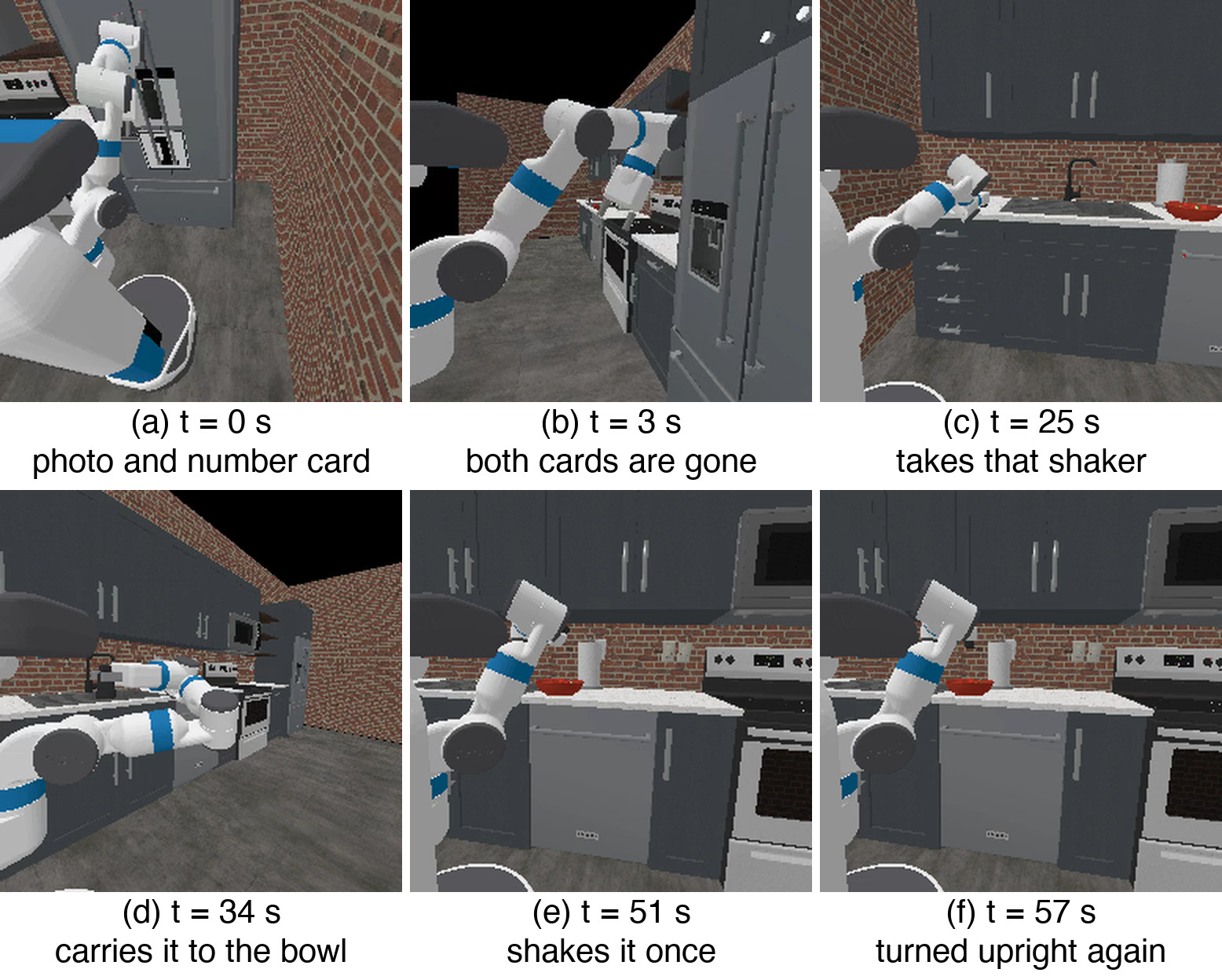}
  \caption{\textbf{SeasonDish.} The initial visual cue specifies both the
    target shaker and the number of shaking actions. In this example, the
    numerical target is one. First released demonstration, right head-mounted
  camera.}
  \label{fig:app-seasondish}
\end{taskfigure}

\taskfield{Scene and objective} A bowl and two shakers of the same model,
a white salt shaker and a black pepper shaker, are placed on the counter
approximately 0.24\,m apart. A photograph of one shaker and a numerical
card are displayed on the refrigerator. The robot begins 0.9\,m from the
refrigerator, facing the cue, and must perform the specified number of
shaking actions using the indicated object.

\taskfield{Language instruction} ``Remember whether the picture on the
fridge shows the white salt shaker or the black pepper shaker, and the
number next to it. After the picture disappears, take that shaker, turn it
upside down over the bowl, shake it as many times as the number showed and
turn it upright again, leaving the other shaker untouched.''

\taskfield{Memory requirements} The photograph and digit are removed after
2.5\,s. The task combines recall of the target object's identity with
retention of a numerical target and tracking of the completed shaking
actions. In 100 scripted evaluation episodes, the pick phase starts
9--15\,s (median 10\,s) after cue removal, and the final upright turn occurs
28--74\,s (median 54\,s) after cue removal.

\taskfield{Initial-state randomization} The target shaker, the left--right
ordering of the shakers, and the digit (1--9) are sampled independently.
The bowl and shaker pair are placed at sampled free counter positions with
random yaw. The base position and yaw vary by up to 5\,cm per horizontal
axis and 0.10\,rad. Initial arm, torso, and head deviations from rest are
bounded by 0.04\,rad, 1\,cm, and 0.05\,rad, respectively.

\taskfield{Success and termination} The robot must execute exactly the
specified number of shakes with the target shaker over the bowl, then
hold it upright and stationary for 0.5\,s. A shake is defined as an angular
excursion of at least $20^\circ$ from the inverted orientation followed
by a return to within $10^\circ$ of that orientation. The shake and final
hold conditions are counted only when the non-target shaker is ungrasped
and within 10\,cm of its initial location. Exceeding the target count or
returning upright before the final shake terminates the episode in failure.
An incorrect initial grasp can be corrected within the remaining episode.
The evaluation horizon is 1{,}656 environment steps (165.6\,s).
The demonstrations were recorded with code version \texttt{1274faf}
(collection profile 28), before the premature-upright termination condition
was added in version \texttt{e0558f5}. The evaluation code (version
\texttt{a4b2e99}) includes this condition, and the success and failure
criteria are otherwise unchanged. The recorded simulator states of all
1{,}164 successful planner executions, including the 1{,}000 released
demonstrations, store the shake counter, its arming flag, and the pose of the
target condiment at every control step. In none of them does the armed
target come within $15^\circ$ of upright before the target count is reached.
The smallest tilt in that state is $141^\circ$, and every execution ends with
exactly the target count. The 100 evaluation seeds were verified with the
same planner version. Their planner executions are not released, so we did
not re-check them against the new condition.

\taskfield{Random-choice reference} Uniformly guessing the target count
has probability $1/9$ of being correct, and a uniform first choice of shaker
has probability $1/2$ of being correct. These probabilities describe
individual decisions, rather than an episode-level success rate, since
an incorrect initial object selection can be corrected.

\taskfield{Scripted planner}
\emph{Cue acquisition and approach.} Observe the cue until removal, then
navigate to the shakers and stop 0.10\,m short of them. Select a side grasp
3\,cm below the top from up to eight arm configurations for which lift,
transport, and pouring motions can be planned.
\emph{Acquisition and transport.} Approach from above, descend, move into the
grasp, close the gripper, and lift by approximately 0.15\,m. Verify the grasp
and that the other shaker has not moved. If necessary, navigate to the bowl
while maintaining a folded arm configuration and an upright shaker.
\emph{Shaking.} Position the shaker above the bowl, rotate it nearly upside
down, and execute the target number of $30^\circ$ wrist excursions toward
the upright orientation and back. Return the shaker upright and hold until
success.

\subsection{\textbf{DepthRecall}}
\label{app:depthrecall}

\begin{taskfigure}
  \centering
  \includegraphics[width=\linewidth]{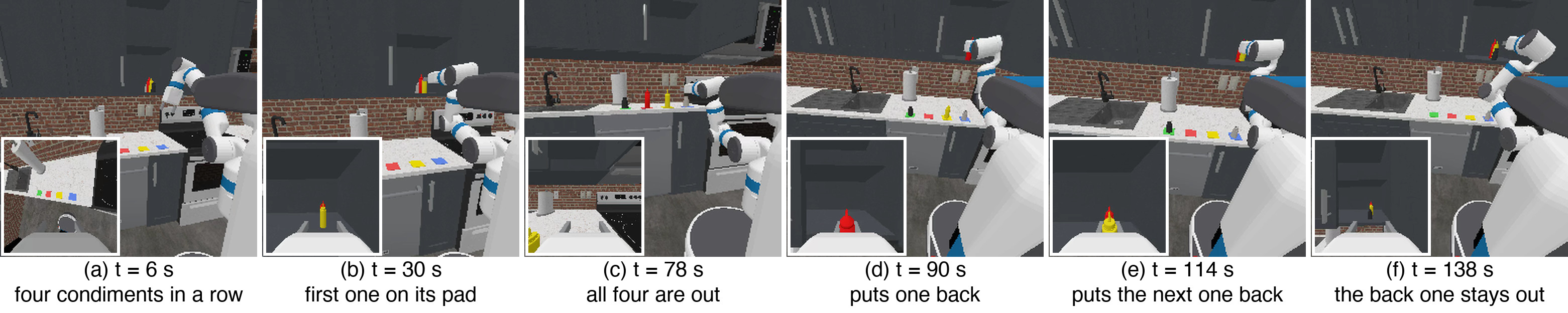}
  \caption{\textbf{DepthRecall.} The robot removes four condiment containers
    from a shelf, retains the initially backmost container, and restores the
  remaining objects to their original positions.}
  \label{fig:app-depthrecall}
\end{taskfigure}

\taskfield{Scene and objective} Four condiment containers stand one behind
another on the shelf of a wall cabinet, in seats 7\,cm apart on a shelf
surface 1.42\,m high: ketchup, a black pepper shaker, a white salt shaker,
and a mustard bottle. They are the RoboCasa objects \texttt{ketchup\_7},
\texttt{shaker\_0}, \texttt{shaker\_1}, and \texttt{ketchup\_8} recolored
yellow, with heights of 16.5, 9.45, 9.45, and 15.0\,cm. The cabinet door is
open at 1.75\,rad from the start. Four colored $9\times9$\,cm pads, one per
container, lie in a line along the front of the counter below, at 0.18\,m
spacing and extending 0.54\,m to one side of the row. The robot moves every container to
its own pad, keeps the container that stood farthest back on its pad, and
returns the other three to their original seats.

\taskfield{Language instruction} ``Take the props out of the cabinet onto the
counter pads (ketchup on the red pad, black pepper shaker on the green pad,
white salt shaker on the blue pad, mustard on the yellow pad), keep the one
from the back, and put the others back where they stood.'' The demonstrations
were recorded with an earlier wording that named the shakers only as pepper
and salt. The released dataset carries the text above.

\taskfield{Memory requirements} The initial front-to-back ordering specifies
both the container to retain and the seat of each container to restore. The
instruction names the pad of each container, so the pads require no memory.
The pad positions are drawn independently of the row order, so the
arrangement on the counter carries no record of the original ordering, and
the emptied shelf has no seat markers. In 100 scripted evaluation episodes,
the restoration of a container begins 18--102\,s (median 54\,s) after its
removal to the pad ends, over the 300 restored containers.

\taskfield{Initial-state randomization} The assignment of containers to the
four seats is a uniform random permutation. The assignment of pads to the four
counter positions is a second permutation, drawn independently. The row and
the pad line shift together by up to 4\,cm along the counter. The robot starts
facing the cabinet about 1.5\,m in front of it, displaced by up to 8\,cm along
the counter relative to the row and by up to 8\,cm across it, with its heading
varied by up to 0.10\,rad. The displacement along the counter uses the same
random draw as the row shift. Initial deviations from the rest configuration
are bounded by 0.04\,rad for the arm, 1\,cm for the torso, and 0.05\,rad for
the head joints. The container orientations and the door angle are fixed.

\taskfield{Success and termination} Success requires every container to
have left its seat and to have been released on its own pad, within 8\,cm of
the pad center. The container that stood farthest back must then rest on its
own pad, and each of the other three must rest within 3.5\,cm of its own
original seat. Resting means released, upright (tilt at most $15^\circ$, with
the base within 1.5\,cm of the supporting surface), and still (linear speed
below 0.05\,m/s and angular speed below 0.2\,rad/s). This arrangement must
persist for 15 control steps (0.75\,s). The order in which the containers
are returned is not checked, although the front seats physically block the
deeper ones. The following violations irreversibly invalidate the episode:
\begin{itemize}
  \item Releasing a container on the pad of another container.
  \item Releasing a container that is being returned in the seat of another
    container.
  \item Returning the container that stood farthest back to any seat.
  \item Two containers resting in one seat.
\end{itemize}
Invalidated episodes continue until the evaluation horizon of 2{,}986
environment steps (298.6\,s). The registered horizon is 5{,}700 control steps
(285\,s).

\taskfield{Random-choice reference} Uniformly guessing the container to retain
and the seats of the other three yields a correct arrangement with probability
$1/24$, and with probability $1/6$ when the retained container is known.

\taskfield{Scripted planner}
\emph{Unloading.} Look at the row and dock once in front of the open cabinet.
For each container, from the front to the back, raise the torso to the shelf,
drive in, grasp the container from the side, lift it by 3\,cm, pull it back
by 6\,cm, and back out. Carry it to its own pad, lower and release it, and
return to the docking line.
\emph{Restoration.} Look at the pads. Return the three containers other than
the one that stood farthest back, filling the deepest empty seat first: grasp
the container on its pad, lift it, drive in, place it in its seat, release
it, and back out.
\emph{Completion.} Hold the robot still for at most 30 control steps until the
arrangement registers as successful.
Each grasp tries three grasp depths at up to three grasp heights and several
torso heights before it is abandoned. A failed planner check aborts the
attempt without a retry.

\subsection{\textbf{Demonstration Collection Statistics}}
\label{app:demonstrations}

The scripted planners are used to collect successful demonstration
trajectories and verify the solvability of held-out evaluation instances.
The collection-attempt statistics are summarized in
Table~\ref{tab:app-planner-stats}. Collection success rates are computed
before filtering and are distinct from VLA evaluation success rates.
Over the six tasks, the planners succeeded in 6{,}722 of 7{,}841 attempts.
The success rate reported in Section~\ref{sec:dataset} is this
pooled rate (85.7\%). The unweighted mean over the tasks is 86.9\%.

\begin{table}[!htbp]
  \centering
  \caption{\textbf{Scripted planner collection statistics.} Attempts are
    training-collection attempts, each on a distinct seed. Accepted
    attempts are successful attempts that also passed the automatic
    acceptance checks described below. The number of released trajectories
  is fixed at 1{,}000 per task.}
  \label{tab:app-planner-stats}
  \small
  \setlength{\tabcolsep}{6pt}
  \renewcommand{\arraystretch}{1.15}
  \begin{tabular}{@{}lrrrrr@{}}
    \hline
    \textbf{Task} & \textbf{Attempts} & \textbf{Successful} &
    \textbf{Success rate (\%)} & \textbf{Accepted} & \textbf{Released} \\
    \hline
    \textsc{TakeItBack} & 1{,}202 & 1{,}079 & 89.8 & 1{,}000 & 1{,}000 \\
    \textsc{SameDrawer} & 1{,}462 & 1{,}167 & 79.8 & 1{,}106 & 1{,}000 \\
    \textsc{CabinetSearch} & 1{,}080 & 1{,}065 & 98.6 & 1{,}062 & 1{,}000 \\
    \textsc{Veggie} & 1{,}557 & 1{,}190 & 76.4 & 1{,}000 & 1{,}000 \\
    \textsc{SeasonDish} & 1{,}440 & 1{,}164 & 80.8 & 1{,}164 & 1{,}000 \\
    \textsc{DepthRecall} & 1{,}100 & 1{,}057 & 96.1 & 1{,}057 & 1{,}000 \\
    \hline
  \end{tabular}
\end{table}

\taskfield{Perturbations} In \textsc{SameDrawer}, \textsc{CabinetSearch},
\textsc{SeasonDish}, and \textsc{DepthRecall}, the planners perturb two
quantities. Selected planning goals, such as approach, docking, hover, and
transfer waypoints, receive independent uniform offsets of up to $\pm5$\,mm
per world axis. In \textsc{CabinetSearch}, the same offsets apply to the
handle grasp point in depth and height and to the can contact point sideways
and in height. During contact-free transfer motions, a Gaussian offset with a
standard deviation of 0.003\,rad is added to each of the seven arm-joint
targets and redrawn every 10 control steps (0.5\,s). Grasping, release, and
other contact phases are executed without this offset. In
\textsc{TakeItBack} and \textsc{Veggie}, selected planner distances are
perturbed uniformly by up to $\pm5\%$ of their nominal values, for example
the 0.50\,m approach standoff in \textsc{Veggie}. The arm-joint targets of
designated motions receive the same Gaussian offset with a standard deviation
of 0.001\,rad, redrawn every 10 control steps. In all tasks, the recorded
actions are the perturbed commands that were executed.

\taskfield{Downsampling and alignment} Each 10\,Hz action combines a pair of
consecutive 20\,Hz control commands. In \textsc{SameDrawer},
\textsc{CabinetSearch}, \textsc{SeasonDish}, and \textsc{DepthRecall}, the
planners issue their commands on a 10\,Hz clock, so the two commands of a pair
are identical, including the binary gripper command (one of two values, $-1$
and $+1$), and gripper switches occur only on even control steps. The 10\,Hz
action therefore equals each command of its pair, and no gripper switch is
lost. In \textsc{TakeItBack} and \textsc{Veggie}, the planners act at
20\,Hz. The two base-velocity components are averaged over each pair of
20\,Hz commands, and observations and the position commands of the arm,
torso, head, and gripper are taken at every second control step. An unpaired
final command is kept. In all tasks, frame $t$ contains the
observation at control step $2t$, immediately before the action of that
frame, and the action covers control steps $2t$ and $2t+1$. Timestamps
advance by 0.1\,s per frame.

\taskfield{Replay validation} In \textsc{SameDrawer}, \textsc{CabinetSearch},
\textsc{SeasonDish}, and \textsc{DepthRecall}, every successful attempt is
replayed twice from a fresh environment reset with the same seed: first with
the recorded 20\,Hz commands, then with the 10\,Hz commands, each held for
two control steps. Both replays must succeed. Observations are rendered from
the simulator states of the second replay, so every released episode is a
successful 10\,Hz execution. For \textsc{TakeItBack} and \textsc{Veggie}, the
exported 10\,Hz actions of 100 episodes per task were replayed from fresh
resets at a 10\,Hz control frequency. The 100 episodes were 20 fixed pilot
episodes and 80 episodes drawn uniformly, all selected before the replay.
All 100 \textsc{TakeItBack} replays succeeded, and 98 of 100 \textsc{Veggie}
replays succeeded. The two failed \textsc{Veggie} replays place the correct
vegetable in the pot and succeed earlier in the replay, but at the final
frame the vegetable's angular speed (0.80 and 0.54\,rad/s) exceeds the
0.5\,rad/s stillness threshold.

\taskfield{Acceptance and selection} In \textsc{SameDrawer},
\textsc{CabinetSearch}, \textsc{SeasonDish}, and \textsc{DepthRecall}, an
accepted attempt passed both replays and the per-batch audits of the
collection pipeline. The released 1{,}000 episodes of \textsc{CabinetSearch},
\textsc{SeasonDish}, and \textsc{DepthRecall} are the first 1{,}000 accepted
episodes in collection order. For \textsc{SameDrawer}, they are the first 250
accepted episodes for each of the four initially open drawers. In
\textsc{SameDrawer}, \textsc{CabinetSearch}, and \textsc{SeasonDish}, the
selection also requires a motion audit: at most 5\% of adjacent 10\,Hz
actions are identical, and no pause exceeds 20 frames. In \textsc{SeasonDish},
the cue period and the final upright hold are excluded from this audit. In
\textsc{TakeItBack} and \textsc{Veggie}, successful attempts were rejected
when the wrist roll or the torso turned by more than $180^\circ$ or when a
collision was detected. Collection stopped once 1{,}000 successes had passed
these checks, and all of them were released.

\clearpage
\section{Baseline Configurations and Training Hyperparameters}
\label{app:hyperparams}

This section reports the architecture-specific settings used in the
experiments. The configurations distinguish model initialization, trainable
components, observation preprocessing, action representation, optimization,
and memory-specific parameters. The prediction horizon and the execution
horizon refer to the number of actions predicted at a policy query and the
number executed before the next query, respectively. A single reported
chunk length denotes the configured prediction length.

All models were trained on four H100 GPUs. The three $\pi_{0.5}$-family runs
share the released RoboMME training stack (JAX~0.5.3, Flax~0.10.2,
Optax~0.2.4); HAMLET runs on GR00T N1.6 under Python~3.12, Torch~2.7.1, and
FlashAttention~2.7.4, and the remaining baselines use their own PyTorch
environments. Evaluation is performed separately in ManiSkill~3.0.0b22 with
Torch~2.7.1 and CPU rendering, at a \qty{20}{\hertz} control rate over
\qty{100}{\hertz} physics, against demonstrations recorded at \qty{10}{\hertz};
scenes are instantiated from the held-out seeds in
\texttt{validation\_seeds.json}. The memoryless $\pi_{0.5}$ baseline is
specified first, and the memory variants are given below either as overrides of
Table~\ref{tab:pi05-config} and Table~\ref{tab:memer-low} or as complete
configurations of their own.

\subsection{\textbf{Memoryless $\pi_{0.5}$ Baseline}}
\label{app:pi05}

\begin{appconfig}{\textbf{Memoryless $\pi_{0.5}$ baseline: model and training configuration.}}{tab:pi05-config}
Backbone & $\pi_{0.5}$ base: PaliGemma 2B vision-language model with a 300M action expert \\
Initialization & \texttt{pi05\_base} orbax checkpoint (12.5 GiB) \\
Trainable components & Language expert, action expert, action input and output adapters \\
Frozen components & SigLIP vision encoder (23 of 23 leaves unchanged in the parameter audit) \\
Observations & Three camera views (two at $256\times256$, one at $128\times128$) resized to $224\times224$; 12D robot state kept in the adapter, without proprioceptive conditioning \\
Action representation & 13D Fetch commands in a 32-channel padded space; the loss reduces over the 13 valid channels only \\
Normalization & $q_{0.01}/q_{0.99}$ quantiles \\
Prediction / execution horizon & 20 / 16 (\qty{2}{\second} at \qty{10}{\hertz}) \\
Optimizer & AdamW, $\beta_1=0.9$, $\beta_2=0.95$, weight decay 0, gradient clipping at 1 \\
Learning rate and schedule & $5\times10^{-5}$, constant after a warmup ratio of 0.05 (4{,}000 steps) \\
EMA decay & 0.999 \\
Batch size & $64 = 16$ samples per device $\times$ 4 H100 GPUs, no gradient accumulation \\
Optimization steps & 80{,}000 ($5.12$ M anchors, approximately one pass over the corpus) \\
Precision and parallelism & bf16, FSDP over four devices \\
Wall-clock time & 22.5 h on four H100 GPUs
\end{appconfig}

\subsection{\textbf{MemER}}
\label{app:memer}

MemER~\citep{memer} uses a high-level vision-language model for subgoal
prediction and a low-level VLA for action generation. The high-level
configuration follows the original implementation, with LoRA adapters
applied without the original weight-merging procedure. The low-level
configuration follows the RoboMME reproduction~\citep{robomme}.

Keyframes are not scored by a learned selector. Subtask boundaries are
annotated in every demonstration, and each subtask type carries a
human-specified first/last/no-frame rule that is applied automatically to all
demonstrations; the resulting frame indices are the keyframe supervision
targets of the high-level model.

\begin{appconfig}{\textbf{MemER: high-level configuration.}}{tab:memer-high}
Backbone & Qwen3-VL-4B \\
Trainable components & LoRA on the LLM, rank $r=16$, scaling $\alpha=32$ \\
Frozen components & Vision encoder and merger \\
Camera preprocessing & Three views stacked vertically into a $192\times576$ image, 108 tokens \\
Temporal input & Frames at $t$, $t-5$, and $t-10$ (\qty{2}{\hertz}) \\
Sampling & One sample per frame \\
Learning rate & $6\times10^{-5}$, cosine schedule \\
Warmup ratio & 0.05 \\
Batch size & $128 = 4$ data-parallel workers $\times$ 4 samples per device
$\times$ 8 gradient-accumulation steps \\
Optimization steps & 1{,}500 \\
Subtask supervision & Scripted event labels and the keyframe-selection rule
\end{appconfig}

\begin{appconfig}{\textbf{MemER: low-level configuration.}}{tab:memer-low}
Backbone & $\pi_{0.5}$-DROID, SigLIP frozen \\
Current observations & Three camera views at $224\times224$, in the left-base, left-wrist, and right-base input slots \\
Proprioceptive input & Disabled \\
Action representation & 13D Fetch commands (arm, head, torso, base, and gripper) \\
Normalization & $q_{0.01}/q_{0.99}$ quantiles \\
Prediction / execution horizon & 20 / 16 \\
Batch size & $128 = 32$ samples per device $\times$ 4 H100 GPUs, no gradient accumulation \\
Optimization steps & 80{,}000 \\
Learning rate & $5\times10^{-5}$, constant after warmup \\
Warmup steps & 10{,}000 \\
Optimizer & AdamW, $\beta_1=0.9$, $\beta_2=0.95$, gradient clipping at 1 \\
Weight decay & 0 ($10^{-10}$ in openpi) \\
EMA decay & 0.999
\end{appconfig}

\subsection{\textbf{FrameSamp+Modul and TokenDrop+Modul}}
\label{app:framesamp-tokendrop}

Both perceptual-memory baselines follow the $\pi_{0.5}$-based
RoboMME architectures~\citep{robomme} and inherit the settings of
Table~\ref{tab:memer-low} (frozen SigLIP, disabled proprioception, EMA 0.999,
gradient clipping at 1, zero weight decay, 16 of 20 predicted actions executed)
with three overrides: both start from the $\pi_{0.5}$ base checkpoint instead of
the DROID checkpoint, use a global batch of 64 instead of 128, and warm up for
4{,}000 instead of 10{,}000 steps, a warmup ratio of 0.05 of the 80{,}000-step
budget. Max pooling follows the original RoboMME paper rather than the mean
pooling of its public release and is applied to the frozen SigLIP patch grid
before the memory tokens are formed: the $16\times16$ patch tokens of one frame
are pooled to $4\times4$ for FrameSamp and to $8\times8$ for TokenDrop, and only
this pooled grid is read by the memory attention, while the current observation
enters the policy with unpooled visual tokens. The policy receives all three
current views and uses a single camera stream for historical features.

\begin{appconfig}{\textbf{FrameSamp+Modul: temporal sampling and optimization.}}{tab:framesamp}
Backbone & $\pi_{0.5}$, SigLIP frozen \\
Learning rate & $5\times10^{-5}$, constant after 4{,}000 warmup steps \\
Batch size & 64 \\
Optimization steps & 80{,}000 \\
Prediction horizon & 20 \\
Frame sampling & 32 frames sampled uniformly from the episode prefix \\
Historical camera & \texttt{left\_base\_camera\_link} \\
Historical image size & $224\times224$ \\
Tokens per frame & 16, using $4\times4$ max pooling \\
Positional encoding & M-RoPE, $768\rightarrow1024$ \\
Memory token dimension & 1{,}024 \\
AdaLN & Identity \\
Memory budget & 512 tokens
\end{appconfig}

\begin{appconfig}{\textbf{TokenDrop+Modul: token selection and optimization.}}{tab:tokendrop}
Backbone & $\pi_{0.5}$, SigLIP frozen \\
Learning rate & $5\times10^{-5}$, constant after 4{,}000 warmup steps \\
Batch size & 64 \\
Optimization steps & 80{,}000 \\
Prediction horizon & 20 \\
Historical camera & \texttt{left\_base\_camera\_link} \\
Tokens per frame & 64 ($8\times8$ over SigLIP features) \\
Memory budget & 512 tokens \\
RGB-difference stride & 8 \\
Selection threshold & $10^{-4}$ \\
Initial frame & Retained in full \\
Selection & Top-512 tokens maintained by a live heap over the episode prefix $0,\ldots,t$
\end{appconfig}

\subsection{\textbf{HAMLET}}
\label{app:hamlet}

HAMLET~\citep{hamlet} is trained in two stages. First, the moment tokens
and memory components are initialized using a reimplementation of the
contrastive objective. Second, the policy is fine-tuned with the GR00T
N1.6-3B backbone, which accommodates the heterogeneous camera resolutions
of DS-Fetch. During the second stage, the language and vision encoders
are frozen, and the moment tokens remain trainable.

\begin{appconfig}{\textbf{HAMLET: contrastive initialization (stage 1).}}{tab:hamlet-tcl}
Trainable modules & \texttt{moment\_to\_repr}, \texttt{backbone.moment\_tokens}, \texttt{memory\_transformer} \\
Anchors & Frames sampled from the training trajectories \\
Positives & Photometric augmentations of the anchor frame (brightness, contrast, blur, noise, small patch occlusion) \\
Negative sampling & A frame from the same trajectory at least one action chunk (40 frames) away \\
Similarity & Cosine similarity of $L_2$-normalised moment representations, obtained by mean-pooling the moment tokens of a frame and passing them through $\mathrm{Linear}(d,d)$, SiLU, and $\mathrm{Linear}(d,d)$ \\
Objective & Cross-entropy over the logits $[\mathrm{sim}(a,p),\mathrm{sim}(a,n)]/\tau$, target index 0 \\
Temperature & $\tau=0.07$ \\
Optimizer and schedule & AdamW, $10^{-4}$ (release default; the paper reports $10^{-5}$), warmup ratio 0.05 with cosine decay \\
Batch size & $64 = 16$ samples per device $\times$ 4 H100 GPUs, no gradient accumulation \\
Optimization steps & Up to 30{,}000 \\
Carried into stage 2 & The four moment-token embeddings $(4\times2048)$, frozen there; the contrastive head is discarded and the memory transformer is re-initialised
\end{appconfig}

\begin{appconfig}{\textbf{HAMLET: policy fine-tuning (stage 2).}}{tab:hamlet-ft}
Backbone & GR00T N1.6-3B \\
Trainable components & Projector, diffusion model, and moment tokens \\
Frozen components & LLM and vision tower \\
Learning rate & $10^{-4}$ \\
Warmup & Ratio 0.05 (3{,}000 steps) \\
Weight decay & $10^{-5}$ \\
Optimizer & AdamW (\texttt{adamw\_torch}), bf16, gradient clipping at 1, ZeRO-2 \\
Batch size & $32 = 8$ samples per device $\times$ 4 H100 GPUs, no gradient accumulation \\
Optimization steps & 60{,}000 \\
Prediction horizon & 40 \\
Temporal sampling & $K=4$, stride 16 \\
Observation window & $[-48,-32,-16,0]$ \\
Moment tokens & 4 \\
Memory layers & 2 \\
Memory configuration & \texttt{cross\_attn}, \texttt{moment\_token} \\
EMA & Disabled \\
Hardware and launcher & Four H100 GPUs, \texttt{torchrun}
\end{appconfig}

\subsection{\textbf{MemoryVLA}}
\label{app:memoryvla}

MemoryVLA~\citep{memoryvla} uses an OpenVLA-7B backbone with a DiT-L
action model and a latent memory bank. The model is adapted to the
13-dimensional DS-Fetch action space and trained using the three camera
views without proprioceptive inputs.

The buffer is specified as a group size and a stream buffer: the group loader
emits 16 consecutive frames per group, and groups are drawn through a
32{,}000-frame shuffle buffer, so temporal adjacency is kept inside a group
while groups are mixed across episodes. Each policy step contributes one
perceptual--cognitive entry, and the bank holds 16 entries per episode.
Retrieval applies two cross-transformer blocks per bank, with query, key, and
value projections, a residual layer normalization, and a $4\times$
feed-forward block, and fuses the retrieved memory with the working memory
through the gate $\sigma(\mathrm{Linear}(2D,D)[x_1;x_2])$ conditioned on a
learned timestep embedding. Consolidation keeps the bank at 16 entries by
merging the most similar adjacent pair of entries instead of discarding the
oldest one.

Action statistics for the per-task min--max normalization are fitted on the
training episodes only: evaluation runs on held-out simulator seeds that never
appear in the corpus, so no evaluation observation enters the ranges.
Optimization uses AdamW with $\beta_1=0.9$, $\beta_2=0.999$, zero weight decay,
gradient clipping at 1, and a constant learning rate of $2\times10^{-5}$ without
warmup.

\begin{appconfig}{\textbf{MemoryVLA: model and training configuration.}}{tab:memoryvla}
Backbone & OpenVLA-7B (Prismatic + DiT-L) \\
Trainable components & All model components \\
Camera configuration & Three views, \texttt{--load\_bi\_wrist} \\
Proprioceptive input & Disabled \\
Action representation & 13D \texttt{FETCH\_JOINT} \\
Normalization & Per-task min--max normalization \\
Prediction / execution horizon & 16 / 8 \\
Buffer configuration & Group 16 / 32{,}000 \\
Batch size & 64 (four H100 GPUs, 16 per GPU) \\
Optimization steps & 160{,}000 \\
Optimizer & AdamW, $\beta_1=0.9$, $\beta_2=0.999$, weight decay 0, clipping at 1, $2\times10^{-5}$ constant without warmup \\
Checkpoint retention & Every 10{,}000 optimization steps, retaining the three most recent checkpoints
\end{appconfig}

\end{document}